%% file: aaai25.tex
\documentclass[letterpaper]{article} % DO NOT CHANGE THIS

\usepackage{geometry}
\usepackage{titlesec}

\usepackage{aaai25}  % DO NOT CHANGE THIS
\usepackage{times}  % DO NOT CHANGE THIS
\usepackage{helvet}  % DO NOT CHANGE THIS
\usepackage{courier}  % DO NOT CHANGE THIS
\usepackage[hyphens]{url}  % DO NOT CHANGE THIS
\usepackage{graphicx} % DO NOT CHANGE THIS
\usepackage{natbib}  % DO NOT CHANGE THIS AND DO NOT ADD ANY OPTIONS TO IT
\usepackage{caption} % DO NOT CHANGE THIS AND DO NOT ADD ANY OPTIONS TO IT
\usepackage{algorithm}
\usepackage{algorithmic}
\usepackage{amsfonts}
\usepackage{amssymb}
\usepackage{amsmath}
\newtheorem{definition}{Definition}

\usepackage{newfloat}
\usepackage{listings}
\DeclareCaptionStyle{ruled}{labelfont=normalfont,labelsep=colon,strut=off} % DO NOT CHANGE THIS
\floatstyle{ruled}
\newfloat{listing}{tb}{lst}{}
\floatname{listing}{Listing}
\title{LayerAct: Advanced Activation Mechanism for Robust Inference of CNNs}
\author{
    Kihyuk Yoon, Chiehyeon Lim
}
\affiliations{
    UNIST, Ulsan, Republic of Korea\\
    kh.yoon@unist.ac.kr, chlim@unist.ac.kr
}

\usepackage{bibentry}
\begin{document}

\maketitle

\begin{abstract}
In this work, we propose a novel activation mechanism called LayerAct for CNNs. This approach is motivated by our theoretical and experimental analyses, which demonstrate that Layer Normalization (LN) can mitigate a limitation of existing activation functions regarding noise robustness. However, LN is known to be disadvantageous in CNNs due to its tendency to make activation outputs homogeneous. The proposed method is designed to be more robust than existing activation functions by reducing the upper bound of influence caused by input shifts without inheriting LN's limitation. We provide analyses and experiments showing that LayerAct functions exhibit superior robustness compared to ElementAct functions. Experimental results on three clean and noisy benchmark datasets for image classification tasks indicate that LayerAct functions outperform other activation functions in handling noisy datasets while achieving superior performance on clean datasets in most cases.
\end{abstract}


\begin{links}
 \link{Code \& Appendix}{https://github.com/KihyukYoon/LayerAct}
 %\link{Appendix}{https://github.com/KihyukYoon/LayerAct}
\end{links}

\section{Introduction}

Most existing activation functions follow \textit{element-level activation (ElementAct)} mechanisms, meaning the functions apply independently to each element of the input. However, we identified a limitation in the method: their robustness varies across samples. This is because their robustness relies on the saturation state, where the activation output for a specific range converges to a certain value. For example, in a layer with ReLU \citep{nair2010rectified}, the elements of the output remain unaffected by input shifts only when they are in the saturation state, specifically when the input and shifted input of the elements are smaller than zero. Consequently, existing activation functions can ensure robustness only when a sufficient number of elements are in the saturation state, not when there are fewer elements in that state.

We found that Layer Normalization (LN; \citet{ba2016layer}) can address this limitation of ElementAct functions. Our theoretical and experimental analyses reveal that placing LN right before the activation (i.e., using the output of LN as the input to the activation) can enhance the robustness of networks. However, previous studies have reported that Convolutional Neural Networks (CNNs) with LN tend to perform poorly on clean datasets, mainly due to LN's tendency to lead to homogeneous activation. These claims are supported by our empirical results on the representation of the last layer in CNNs with LN. Therefore, our goal is to develop a method that captures the noise-robustness benefits of LN without inheriting its homogeneity limitations.

To achieve this goal, we propose a novel activation method, the Layer-level Activation (LayerAct) mechanism. Unlike the ElementAct mechanism, our proposed method utilizes layer-direction normalization for activation, but the activation output is far different from that of ElementAct with LN. This enables LayerAct to absorb LN's robustness benefit for activation while avoiding its homogeneity limitation. Additionally, LayerAct functions are not subject to the trade-off issue between two important properties of activation, one-side saturation and zero-like activation mean \cite{clevert2015fast, qiu2018frelu}, which are faced by ElementAct functions (for details, see Appendix A). These two benefits of LayerAct functions, enhanced robustness and freedom from the trade-off problem, underscore their potential to outperform existing ElementAct functions on both noisy and clean datasets.

Experimental analysis with the MNIST dataset demonstrates that LayerAct functions have the following properties: (1) the mean activation of LayerAct functions is zero-like, and (2) the output fluctuation due to noisy input is smaller with these functions than with ElementAct functions. We compared the performance of the residual networks (ResNets; \citet{he2016deep}) with LayerAct functions to those with other ElementAct functions on three image classification tasks. 

Our contributions can be summarized as follows:

\begin{itemize}
\item We identify a previously unrecognized limitation in existing activation mechanisms, namely a large variance of noise-robustness across samples.
\item We present analysis and experimental evidence showing that while LN can address this limitation, it also causes the final layer representation in CNNs to be less distinguishable between samples with different labels.
\item We introduce LayerAct that exhibit greater robustness than existing ElementAct functions, without the homogeneity limitation of LN.
\item We empirically evaluate LayerAct functions, and they show superior performance on both clean and noisy datasets in most cases.
\end{itemize}
%%%%%%%%%%%%%%%%%%%%%%%%%%%%%%%%%%%%%%%%%%%%%%%%%%%%%%%%%%%%%%%%%
%%%%%%%%%%%%%%%%%%%%%%%%%%%%%%%%%%%%%%%%%%%%%%%%%%%%%%%%%%%%%%%%%
%%%%%%%%%%%%%%%%%%%%%%%%%%%%%%%%%%%%%%%%%%%%%%%%%%%%%%%%%%%%%%%%%

\section{Backgrounds}
\label{section_background}

In this section, we provide the background for the analyses presented in the remainder of this paper, including the definition of \textit{activation scale} and a review of normalizations.

%%%%%%%%%%%%%%%%%%%%%%%%%%%%%%%%%%%%%%%%%%%%%%%%%%%%%%%%%%%%%%%%%

\subsection{Activation Scale}
\label{subsection_activation_scale_function}

In some activation functions, a function bounded between one and zero characterizes their non-linearity. We define this function, denoted as $s$, and its output as the \textit{activation scale function} and \textit{activation scale}, respectively.

Consider a layer in a multi-layer perceptron with linear projection and an activation function. Given a $r$-dimensional vector $x=(x_{1}, ..., x_{r})^{T}$, a weight matrix $W \in \mathbb{R}^{r\times D}$, where d is the dimension of activation input (e.g., $D=C \times H \times W$ for image), and a non-linear activation function $f$, the activation output is $a=f\left(y\right)$, where $y=W^{T}x$. With an activation scale function $s$, the activation output and gradient of forward and backward passes are:
\begin{equation}
\label{eq_af_forward_and_backward}
    a_{i}=y_{i}s\left(y_{i}\right), 
    \quad
    \frac{\partial a_{i}}{\partial y_{i}}
    = s\left(y_{i}\right) + y_{i}\frac{\partial s\left(y_{i}\right)}{\partial y_{i}},
\end{equation}
where $s$ is increasing and $s\left(y_{i}\right)>0$ if $y_{i}>0$. For example, the activation scale function for the $i^{th}$ element in the SiLU \citep{elfwing2018sigmoid} is presented as follows:
\begin{equation}
\label{eq2}
\begin{split}
    &s^{SiLU}\left( y_{i}\right) = sigmoid\left(y_{i}\right) = \frac{1}{1+e^{-y_{i}}},
\end{split}
\end{equation}
where $sigmoid$ and $s^{SiLU}$ represent the Logistic Sigmoid function, and the activation scale functions of SiLU, respectively. For further discussion, we define a specific activation scale function $s$. 
\begin{definition}
\label{def_activation_scale_function}
    \textbf{\upshape (Activation Scale Function)}
    The activation scale function $s$ is an increasing Lipschitz continuous function that is bounded between zero and one: 
    \begin{equation}
    \label{ep_asf_def}
        s \left(0\right)=1/2,
        \;
        \left\vert s\left(a\right) - s\left(b\right) \right\vert 
        \leq K\left\vert a - b\right\vert
        \quad \forall a, b \in \mathbb{R}.
    \end{equation}
\end{definition}

For example, the activation scale of SiLU satisfies this, as the activation scale is $sigmoid\left(y_{i}\right)$. Meanwhile, the activation scale also determines the saturation state of the activation functions, which occurs when $s\left(y_{i}\right)\simeq0$.

In conclusion, the activation scale function plays a crucial role in providing non-linearity during the forward pass, controlling the gradient during the backward pass, and determining the saturation state of the activation function.

%%%%%%%%%%%%%%%%%%%%%%%%%%%%%%%%%%%%%%%%%%%%%%%%%%%%%%%%%%%%%%%%%

\subsection{Revisiting Normalizations in CNNs}

Normalization layers re-scale and re-center both the input and the gradient during forward and backward propagation, respectively. The normalization output $n_{i}$ from the normalization input $y_{i}$ is:
\begin{equation}
\label{eq_norm}
\begin{split}
    n_{i} = \gamma_{i} \cdot \frac{y_{i}-\mu_{y}}{\sigma_{y}}+\beta_{i},
\end{split}
\end{equation}
where $\gamma_{i}$ and $\beta_{i}$ are the learnable parameters of the affine function, $D$ is the normalizing dimension, and $\mu_{y}$ and $\sigma_{y}$ are:
\begin{equation}
\label{eq_norm_mean_var}
\begin{split}
    \mu_{y} = \frac{1}{D}\sum^{D}_{i=1}y_{i}, 
    \quad 
    \sigma_{y} = \sqrt{\frac{1}{D}\sum^{D}_{i=1}\left(y_{i}-\mu_{y}\right)^{2}},
\end{split}
\end{equation}
the mean and variance of the input $y$, respectively. Normalization layers can stabilize the network training by re-scaling and re-centering as shown in Equation \ref{eq_norm}. The dimension $D$ varies according to the normalization methods, $D=N \times H \times W$ for Batch Normalization (BN; \citet{ioffe2015Batch}) and $D=C \times H \times W$ for LN, where $N$, $C$, $H$, and $W$ denote the size of the batch size, channel, height, and width of the image dataset. 

BN has achieved great success in computer vision tasks within CNN-based networks and remains dominant in the field. In contrast, LN is less preferred compared to BN, largely due to the inferior performance of networks using LN. Previous studies have attributed this underperformance to LN's tendency to produce homogeneous outputs \cite{Lubana2021Beyond, labatie2021proxy}.

%%%%%%%%%%%%%%%%%%%%%%%%%%%%%%%%%%%%%%%%%%%%%%%%%%%%%%%%%%%%%%%%%
%%%%%%%%%%%%%%%%%%%%%%%%%%%%%%%%%%%%%%%%%%%%%%%%%%%%%%%%%%%%%%%%%
%%%%%%%%%%%%%%%%%%%%%%%%%%%%%%%%%%%%%%%%%%%%%%%%%%%%%%%%%%%%%%%%%

\section{Problems Analysis and Research Motivation}

In this section, we present our motivation, specifically, we focus on: (1) an analysis of the limitations of existing activation functions, and (2) theoretical and empirical analyses of the advantages and disadvantages of LN in CNNs. 

%%%%%%%%%%%%%%%%%%%%%%%%%%%%%%%%%%%%%%%%%%%%%%%%%%%%%%%%%%%%%%%%%

\subsection{Large Variance of Activation Robustness}
\label{subsection_large_variance}

To analyze the robustness, we define \textit{activation fluctuation} that can represent the influence of the shift of inputs. 
\begin{definition}
\textbf{\upshape (Activation Fluctuation)}
\label{def_activation_fluctuation}
    Let $\epsilon=(\epsilon_{1}, \epsilon_{2}, ...,$ $ \epsilon_{D})^{T}$ be the input shift vector, which is independent of the input vector $y$. We define activation fluctuation as $\left\Vert f\left(y+\epsilon\right)-f\left(y\right)\right\Vert \leq c$, where the constant $c$ is the upper bound of activation fluctuation. 
\end{definition}

The lower the upper bound $c$ is, the lower the variance of robustness across samples. We can define the activation fluctuation of ElementAct functions.
\begin{definition}
\label{def_activation_fluctuation_element}
    \textbf{\upshape (Activation Fluctuation of Element-level \ Activation Functions)}
    Let $\hat{y}_{i} = y_{i} + \epsilon_{i}$ be the input with a shift. The activation fluctuation of an ElementAct function $f$ is given by:
    \begin{equation}
    \begin{split}
        \left\Vert f\left(\hat{y}\right) - f\left(y\right) \right\Vert
        &= 
        \sum^{D}_{i=1}
        \left\vert 
            \hat{y}_{i} s\left(\hat{y}_{i}\right)
            - y_{i}s\left(y_{i}\right)
        \right\vert 
        \\
        &= 
        \sum^{D}_{i=1}
        \left\vert 
            y_{i} \left(s\left(\hat{y}_{i}\right) - s\left(y_{i}\right)\right)
            + \epsilon_{i} s\left(\hat{y}_{i}\right)
        \right\vert.        
    \end{split}
    \end{equation} 
\end{definition}

A sample will exhibit a small $\left\Vert f\left(\hat{y}\right) - f\left(y\right) \right\Vert$ if a sufficient number of its elements are in a saturation state. However, ElementAct functions do not ensure that all samples have a sufficient number of elements in the saturation state. More specifically, the activation fluctuation is upper-bounded when not all elements are in the saturation state, where $y_{i}>0$ for all $i$:
\begin{equation}
\label{eq_fluctuation_element_upper}
    \begin{split}
        \left\Vert f\left(\hat{y}\right) - f\left(y\right) \right\Vert
        \leq
        \sum^{D}_{i=1} 
        \left(
            y_{i}  
            \left\vert s\left(\hat{y}_{i}\right) - s\left(y_{i}\right) \right\vert
            + \left\vert \epsilon_{i} \right\vert \cdot s\left(\hat{y}_{i}\right)
        \right).
    \end{split}
\end{equation}  

Considering Definition \ref{def_activation_scale_function}, the upper bound of $\left\Vert s\left(\hat{y}\right) - s\left(y\right) \right\Vert$ and $\left\Vert s\left(\hat{y}\right)\right\Vert$ of ElementAct functions are given by the following, respectively:
\begin{equation}
\label{eq_s_upper_element}
    \left\Vert s\left(\hat{y}\right) - s\left(y\right) \right\Vert
    \leq \sum^{D}_{i=1} K \left\vert \epsilon_{i} \right\vert,
    \quad
    \left\Vert s\left(\hat{y}_{i}\right) \right\Vert \leq d.
\end{equation}

Equation \ref{eq_fluctuation_element_upper} demonstrates that activation scale is closely related to the activation fluctuation. Samples with large $\left\Vert s\left(\hat{y}\right) - s\left(y\right) \right\Vert$ and $\left\Vert s\left(\hat{y}\right)\right\Vert$ are not robust to the input shift $\epsilon$. Thus, a method that has a lower upper bound for $\left\Vert s\left(\hat{*}\right) - s\left(*\right) \right\Vert$ and $\left\Vert s\left(\hat{*}\right)\right\Vert$ will reduce the upper bound of activation fluctuation, resulting in a low variance of robustness across samples.

%%%%%%%%%%%%%%%%%%%%%%%%%%%%%%%%%%%%%%%%%%%%%%%%%%%%%%%%%%%%%%%%%

\subsection{Empirical Analysis on Normalizations in CNNs}
\label{subsection_LN_activation}

To empirically analyze the impact of BN and LN in CNNs, we conducted a series of experiments using the CIFAR10 \cite{krizhevsky2009learning} dataset with the ResNet20 architecture. In the original ResNet20, BN layers are positioned immediately before the activation layers. For our experiments, we designed four variants of ResNet20: (1) eliminating BN layers entirely (ResNet20-None), (2) replacing BN layers with LN layers (ResNet20-LN), (3) adding LN layers before the BN layers (ResNet20-(LN, BN)), and (4) adding LN layers after the BN layers (ResNet20-(BN, LN)). 

\begin{table}[t]
\begin{center}
\begin{tabular}{cccccc}
Data & BN & None & LN & BN, LN & LN, BN \\
\hline \hline 
C10 & 91.29  & 88.51 & 88.24 & 90.65 & 89.73\\
C10-C & 69.92 & 67.43 & 74.5 & 74.34 & 72.55 \\
\hline 
C10 & 91.45 & 88.29 & 87.53 & 89.74 & 90.17 \\
C10-C & 70.12 & 68.94 & 74.96 & 73.69 & 72.47 \\
\hline 
\end{tabular}
\caption{Classification performance of original ResNet20 and three variants of ResNet20 with ReLU (upper) and SiLU (lower) on CIFAR10 (C10) and CIFAR10-C (C10-C).}
\label{table_motivation_cifar_resnet}
\end{center}
\end{table}

Table \ref{table_motivation_cifar_resnet} demonstrates the performance of the ResNet20 variants with different normalization on both CIFAR10 and CIFAR10-C \cite{hendrycks2019robustness}. The networks were trained on CIFAR10. The CIFAR10-C dataset serves as a noise-robustness benchmark, incorporating 19 out-of-distribution (OOD) corruptions of CIFAR10. This dataset includes a total of 19 distinct corruptions, each with five levels of severity, organized into five categories: noise, blur, digital, weather, and extra. We evaluated the models using two activation functions: ReLU and SiLU. The table reports the average mean accuracy over 30 runs, with "Norm" and "Act" denoting the normalization and activation, respectively. The experimental results show that the use of LN alone leads to inferior performance on the clean dataset, CIFAR10. Specifically, networks with LN alone showed inferior results compared to those without normalization, which aligns with findings from previous studies.

LN and other batch-free normalizations have been shown to block the negative effect of BN on robustness of networks \cite{huang2022delving}. Interestingly, our experiments reveal a detailed robustness advantage of LN, which can lead to robust activation, and this has not been emphasized in previous studies. On the noisy dataset, CIFAR10-C, the networks incorporating LN demonstrated better performance than those without LN. Notably, when LN is applied immediately before the activation layers, as in ResNet20-LN and ResNet20-(BN, LN), these networks outperformed their counterparts, ResNet20-None and ResNet20-(LN, BN), on noisy datasets, despite underperforming on the clean dataset. This suggests that LN may provide a robustness benefit, which is closely related to activation. 

%%%%%%%%%%%%%%%%%%%%%%%%%%%%%%%%%%%%%%%%%%%%%%%%%%%%%%%%%%%%%%%%%

\subsection{LN and Activation Robustness}
\label{subsection_benefit_of_LN}

Based on the experimental result from the experimental analysis, we analyzed the relationship between LN and activation robustness. Here, we only consider LN without an affine function. In this case, the normalized output $n_{i}$, the normalized output with a shift $\hat{n}_{i}$, the activation output $a$, and activation output with shift $\hat{a}$ are as follows:
\begin{equation}
\label{eq_ln_noisy}
\begin{split}
    n_{i}
    = 
    \frac{y_{i}-\mu_{y}}{\sqrt{\sigma^{2}_{y}+\alpha}}, 
    \quad
    \hat{n}_{i} 
    =
    \frac{\hat{y}_{i}-\hat{\mu}_{y}}{\sqrt{\hat{\sigma}^{2}_{y}+\alpha}},
    \\
    a_{i}
    =
    n_{i} s \left( n_{i} \right),    
    \quad
    \hat{a}_{i}
    =
    \hat{n}_{i} s \left( \hat{n}_{i} \right),
\end{split}
\end{equation}
where $\hat{\mu}_{y}$ and $\hat{\sigma}_{y}^{2}$ are the layer-direction mean and variance of $\hat{y}$. The activation fluctuation using Equations \ref{eq_norm} and \ref{eq_ln_noisy} are:

\begin{definition}
\textbf{\upshape (Activation Fluctuation of ElementAct Functions with LN)}
\label{def_activation_fluctuation_LN}
    Let $\epsilon_{i}$, $\hat{y}_{i}$, and $\hat{n}_{i}$ of the $i^{th}$ normalization layer be the shift of the input, the input with the shift, and the output from the input. The activation fluctuation of an ElementAct function $f$ is:
    \begin{equation}
    \label{eq_definition_activation_fluctuation_LN}
    \begin{split}
        \left\Vert f\left(\hat{n}\right) - f\left(n\right) \right\Vert
        =& 
        \sum^{D}_{i=1}
        \left\vert 
            \hat{n}_{i} s\left(\hat{n}_{i}\right)
            - n_{i}s\left(n_{i}\right)
        \right\vert 
        \allowdisplaybreaks
        \\
        \lesssim&
        \sum^{D}_{i=1}
        \left(
            \left\vert n_{i} \right\vert \cdot
            \left\vert
                s\left(\hat{n}_{i}\right)-s\left(n_{i}\right)
            \right\vert
        \right) 
        \allowdisplaybreaks
        \\
        &+
        \sum^{D}_{i=1}
        \left(
        \frac{\left\vert\epsilon - \mu_{\epsilon}\right\vert \cdot s\left(\hat{n}_{i}\right)}{\sqrt{\sigma_{y}^{2}+\alpha}}
        \right), 
    \end{split}
    \end{equation}
where $\sqrt{\sigma_{y}^{2}+\alpha+\sigma_{\epsilon}^{2}} \approx \sqrt{\sigma_{y}^{2}+\alpha}$ and $\sigma_{y}\gg\sigma_{\epsilon}$. 
\end{definition}

For the detailed derivation process of the equations, see Appendix B. In the previous section, we discussed that reducing the magnitudes of the terms of the activation scale effectively decreases the upper bound of activation fluctuation, thereby reducing the variance of activation robustness across samples. Considering Definitions \ref{def_activation_scale_function} and \ref{def_activation_fluctuation_LN}, the upper bound of $\left\Vert s\left(\hat{n}\right) - s\left(n\right) \right\Vert$ and $\left\Vert s\left(\hat{n}\right) \right\Vert$ is given as follows:
\begin{equation}
\label{eq_s_upper_LN}
        \left\Vert s\left(\hat{n}\right) - s\left(n\right) \right\Vert 
        <
        \sum^{D}_{i=1} 
        \frac{K\left\vert\epsilon_{i} -\mu_{\epsilon} \right\vert}{\sqrt{\sigma_{y}^{2}+\alpha}}, 
        \quad
        \left\Vert s\left(\hat{n}_{i}\right) \right\Vert \ll d,
\end{equation} 
which are lower than those of ElementAct functions without normalization (see Equation \ref{eq_s_upper_element}), specifically considering that $\frac{\left\vert\epsilon_{i} -\mu_{\epsilon} \right\vert} {\sqrt{\sigma_{y}^{2}+\alpha}}<1$, where $\sigma_{y}\gg\sigma_{\epsilon}$. This reveals that LN can improve activation robustness not only by re-scaling and re-centering, but also by affecting the activation scale. The experimental results in Table \ref{table_motivation_cifar_resnet} support our analysis, demonstrating that networks with LN are much more robust compared to those without normalization.

%%%%%%%%%%%%%%%%%%%%%%%%%%%%%%%%%%%%%%%%%%%%%%%%%%%%%%%%%%%%%%%%%

\begin{figure}[t]
\centering
\includegraphics[width=0.49\columnwidth]{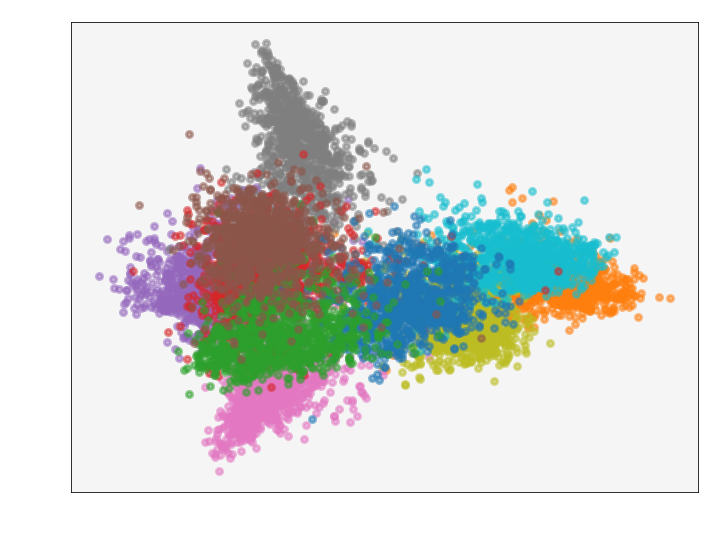}
\includegraphics[width=0.49\columnwidth]{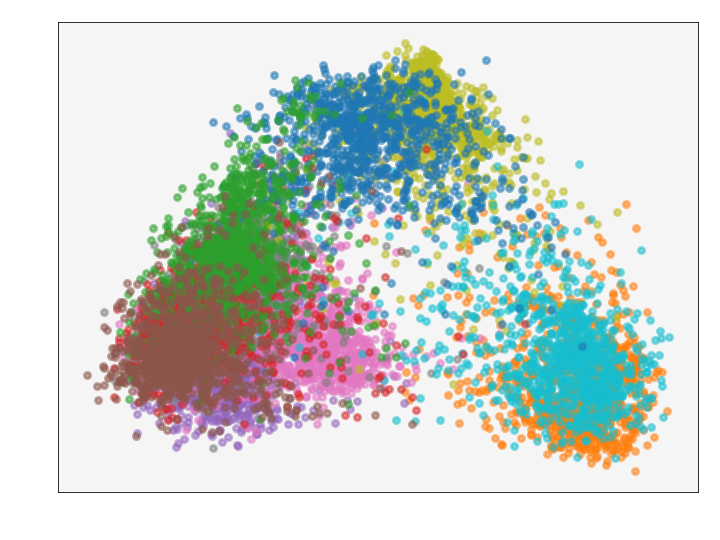}
\caption{Plots of the last layer representation in ResNet20 with BN (left), and LN (right).}
\label{fig_iso_BN_LN}
\end{figure}

\subsection{Homogeneity Limitation of LN across Labels}

It is well-known that LN produces homogeneous outputs across samples by normalizing the mean and variance of the input. However, this homogeneity extends beyond its output: LN causes activations to become similar across samples. Specifically, when LN is placed right before the activation function, producing outputs $n s\left(n\right)$, it loses the mean and variance statistics of input $y$. Here, we provide experimental results showing that LN produces homogeneous representations across samples, even those with different labels.

Figure \ref{fig_iso_BN_LN} illustrates the last layer representation in ResNet20 with BN and LN on CIFAR10, visualized using Isomap embedding \cite{Tenenbaum2000AGlobal}. The figure demonstrates that the representation of the network with LN is less distinguishable between data points with different labels compared to the network with BN, which provides evidence of how LN leads to inferior performance on clean datasets. This reduced distinguishability is due to the similar output statistics of LN across samples. The variances of the output mean and variance in networks with LN are 0.0001 and 0.00072, respectively, which are substantially lower than those observed with BN (0.00073 and 0.00803, respectively). These experimental results align with previous studies, suggesting that LN's tendency to produce homogeneous outputs is a key limitation.

\subsection{Summary}

In this section, we presented analyses and empirical results highlighting that existing activation functions have a limitation regarding noise robustness, which LN can address. However, LN is not a preferred normalization method for CNNs due to its homogeneity limitation. Motivated by this, our goal is to develop a method that captures the robustness benefit of LN without inheriting its limitation.

%%%%%%%%%%%%%%%%%%%%%%%%%%%%%%%%%%%%%%%%%%%%%%%%%%%%%%%%%%%%%%%%%
%%%%%%%%%%%%%%%%%%%%%%%%%%%%%%%%%%%%%%%%%%%%%%%%%%%%%%%%%%%%%%%%%
%%%%%%%%%%%%%%%%%%%%%%%%%%%%%%%%%%%%%%%%%%%%%%%%%%%%%%%%%%%%%%%%%

\begin{figure}[tb]
\centering
\includegraphics[width=0.49\columnwidth]{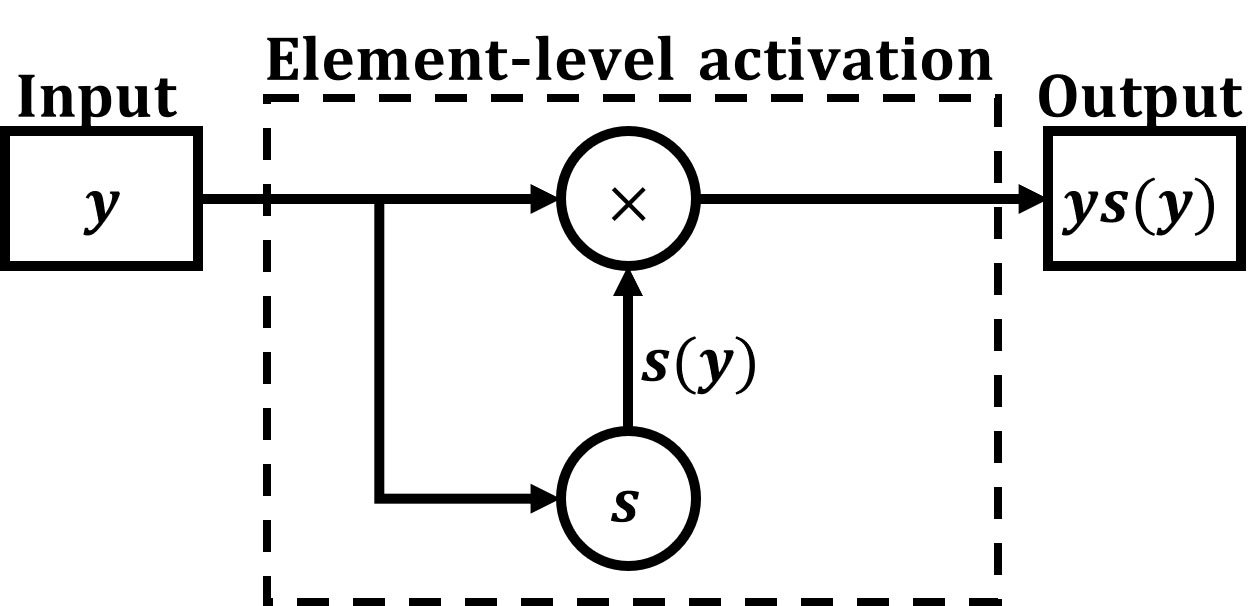} 
\includegraphics[width=0.49\columnwidth]{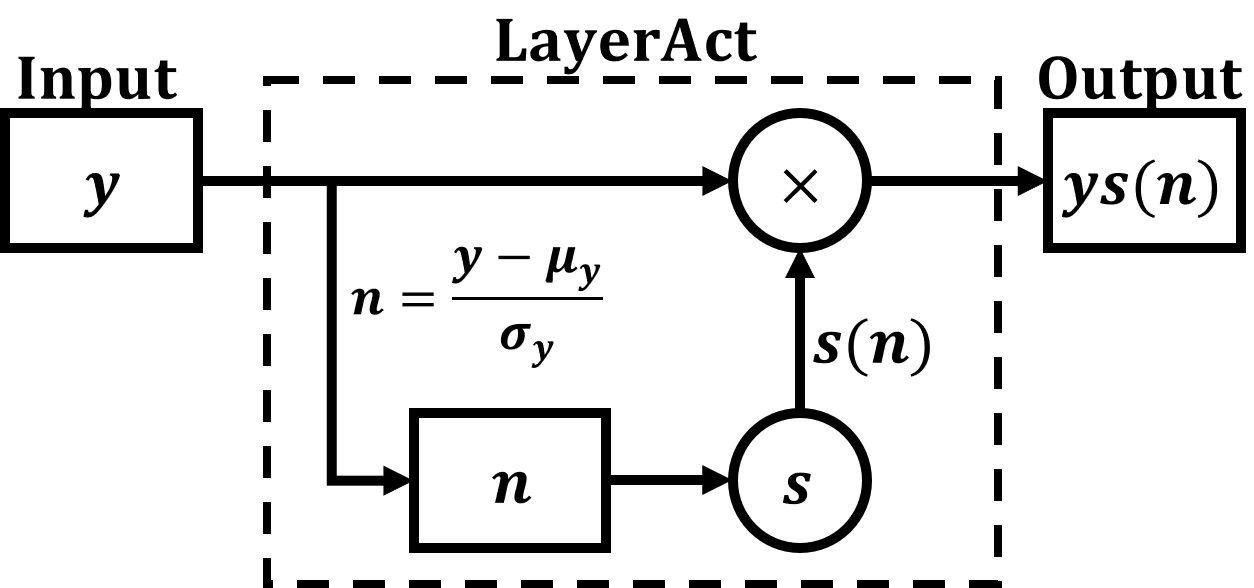} \\
\caption{The mechanisms of the ElementAct (left) and proposed layer-level activation (right).}
\label{fig_element_vs_layeract}
\end{figure}

\section{Layer-level Activation}

In this section, we introduce LayerAct mechanism, a novel approach that operates differently from existing ElementAct functions, as illustrated in Figure \ref{fig_element_vs_layeract}.

%%%%%%%%%%%%%%%%%%%%%%%%%%%%%%%%%%%%%%%%%%%%%%%%%%%%%%%%%%%%%%%%%

\subsection{LayerAct Mechanism}
\label{subsection_LayerAct}

A function that follows LayerAct mechanism is defined as the product of the input $y_{i}$ and the activation scale $s(n_{i})$, which uses the layer-normalized input $n_{i}$. The forward and backward pass of a LayerAct function are given by:

\begin{equation}
\label{eq_layeract_forward}
    a_{i}=y_{i} s\left(n_{i}\right), \quad
    n_{i}=\frac{\left(y_{i}-\mu_{y}\right)}{\sqrt{\sigma_{y}^{2}+\alpha}},
\end{equation}

\begin{equation}
\label{eq_layeract_backward}
\begin{split}
    \frac{\partial \mathcal{L}}{\partial \mu_{y}} =&
    \sum^{D}_{i=1} \frac{\partial \mathcal{L}}{\partial a_{i}}
    \cdot \frac{\partial s\left( n_{i}\right)}{\partial n_{i}}
    \cdot \frac{-y_{i}}{\sqrt{\sigma_{y}^{2}+\alpha}},
    \\
    \frac{\partial \mathcal{L}}{\partial \sigma_{y}^{2}} =&
    \sum^{D}_{i=1} \frac{\partial \mathcal{L}}{\partial a_{i}}
    \cdot \frac{\partial s\left( n_{i}\right)}{\partial n_{i}}
    \cdot \frac{- y_{i} \cdot n_{i} }{2\left(\sigma_{y}^{2}+\alpha\right)},
    \\
    \frac{\partial \mathcal{L}}{\partial y_{i}} =&
    \frac{\partial \mathcal{L}}{\partial a_{i}} s\left(n_{i}\right)
    + \frac{\partial \mathcal{L}}{\partial a_{i}}
    \cdot \frac{\partial s\left( n_{i}\right)}{\partial n_{i}}        
    \cdot \frac{y_{i}}{\sqrt{\sigma_{y}^{2}+\alpha}} 
    \\
    &+ \frac{1}{D} \cdot \frac{\partial \mathcal{L}}{\partial \mu_{y}}
    + \frac{2\left(y_{i}-\mu_{y}\right)}{D} \cdot \frac{\partial \mathcal{L}}{\partial \sigma_{y}^{2}},
\end{split}
\end{equation}
respectively, where  $\alpha>0$ is a constant for stability, and $\mu_{y}$ and $\sigma_{y}^{2}$ are layer-direction mean and variance, respectively. 

For the stability of learning and inference, ensuring the continuity of activation outputs across the entire space is essential. While traditional ElementAct functions, such as ReLU, LReLU \citep{maas2013rectifier}, and PReLU \citep{he2015delving}, do not require a continuous activation scale at zero (given that the activation output $y_{i}s\left(y_{i}\right)$ remains continuous at zero), LayerAct functions need special consideration, as the activation output $y_{i}s\left(n_{i}\right)$ is discontinuous if the activation scale function is not continuous.

Given such requirements for the activation scale functions, functions that satisfy Definition \ref{def_activation_scale_function} are suitable for use as LayerAct functions. In this work, we propose two basic LayerAct functions, LA-SiLU and LA-HardSiLU, which utilize the Sigmoid and HardSigmoid functions as activation scale functions, respectively. The activation outputs of these functions are as follows:
\begin{equation}
\label{eq_layeract_functions}
\begin{split}
    a^{LA\mbox{-}SiLU} =& \frac{y_{i}}{1+e^{-n_{i}}},
    \\
    a^{LA\mbox{-}HardSiLU} =&
    \textit{min}\left(y_{i}, \textit{max}\left(\frac{y_{i}\cdot n_{i}}{6} + \frac{y_{i}}{2}, 0\right)\right).
\end{split}
\end{equation}
For the HardSigmoid of LA-HardSiLU, we used the function of \citet{Howard2019Searching}, which is a good approximation of Sigmoid. 

%%%%%%%%%%%%%%%%%%%%%%%%%%%%%%%%%%%%%%%%%%%%%%%%%%%%%%%%%%%%%%%%%

\subsection{Benefits and Properties of LayerAct}
\label{subsection_noiserobustness_layeract}

\subsubsection{Diverse Activation Output}

We discussed that LN tends to produce homogeneous activation outputs, as the mean and variance of $n_{i}$ are similar across all samples. In contrast, LayerAct functions are designed to preserve these diverse representations by transferring the statistics of the activation input $y$ to the activation output $a$. This difference is from the different activation outputs of activation with LN and LayerAct, $n_{i}s\left(n_{i}\right)$ and $y_{i}s\left(n_{i}\right)$, respectively. Due its unique output, LayerAct can maintain the input statistics within the activation outputs, resulting in much more diverse representations. For details, see Appendices C and D.

\subsubsection{Noise-robustness of LayerAct}

From here, we begin by establishing that the activation fluctuation of LayerAct is also related to the two terms of the activation scale function, $\left\Vert s\left(\hat{*}\right) - s\left(*\right) \right\Vert$ and $\left\Vert s\left(\hat{*}\right)\right\Vert$, as outlined in previous sections. Subsequently, we show that these two terms for LayerAct are bound to be lower than those of ElementAct. We can define the activation fluctuation of LayerAct as follows.
\begin{definition}
\textbf{\upshape (Activation Fluctuation of LayerAct Functions)}
\label{def_activation_fluctuation_layeract}
    The activation fluctuation of a LayerAct activation function $g$,
    where $\hat{n}_{i}=\left(\hat{y}_{i}-\mu_{\hat{y}}\right)/\sigma_{\hat{y}}$ denotes the $i^{th}$ noisy normalized input, is defined as:
    \begin{equation}
    \begin{split}
        \left\Vert g\left(\hat{y}\right) - g\left(y\right) \right\Vert
        =& 
        \sum^{D}_{i=1}
        \left\vert 
            \hat{y}_{i} s\left(\hat{n}_{i}\right)
            - y_{i}s\left(n_{i}\right)
        \right\vert 
        \\
        =&
        \sum^{D}_{i=1}
        \left\vert 
            y_{i} \left(s\left(\hat{n}_{i}\right) - s\left(n_{i}\right)\right)
            + \epsilon_{i} s\left(\hat{n}_{i}\right)
        \right\vert.
    \end{split}
    \end{equation} 
\end{definition}

This definition enables us to establish an upper bound for the activation fluctuation of LayerAct functions as follows: 
\begin{equation}
\label{eq_fluctuation_layeract_upper}
    \left\Vert g\left(\hat{y}\right) - g\left(y\right) \right\Vert
    \leq
    \sum^{D}_{i=1} 
    \left(
         \left\vert y_{i} \right\vert
        \left\vert s\left(\hat{n}_{i}\right) - s\left(n_{i}\right) \right\vert
        + \left\vert \epsilon_{i} \right\vert s\left(\hat{n}_{i}\right)
    \right).
\end{equation}

The terms of LayerAct's activation scale functions, $\left\Vert s\left(\hat{n}\right) - s\left(n\right) \right\Vert$ and $\left\Vert s\left(\hat{n}\right)\right\Vert$, align precisely with those of the activation with LN, as indicated in Equation \ref{eq_s_upper_LN}. Comparing Equations \ref{eq_s_upper_element} and \ref{eq_s_upper_LN} reveals that LayerAct functions have a lower upper bound of activation fluctuation, similar to activation with LN. This reduction is achieved without the direct re-centering and re-scaling of the activation output. Therefore, it suggests that networks with LayerAct are likely to exhibit improved robustness during the forward pass.

\subsubsection{Addressing the Trade-off} 

LayerAct can address another limitation of existing activation functions, namely a trade-off between two important properties of activation: one-side saturation and zero-like activation mean. The key distinction in saturation between the element-level and LayerAct functions is that element-level functions have a fixed saturation state at a certain point of activation output, whereas the saturation state of LayerAct functions depends on layer-dimension normalized inputs. Consequently, LayerAct functions can still reach a saturation state where $s\left(n_{i}\right)\simeq0$, yet without constraining the activation output space, enabling larger negative outputs (e.g., consider a layer where $\mu_{y}\ll0$). For details of the trade-off, see Appendix E.

\subsection{Relationship between Normalization} 

LayerAct functions have unique benefits that ElementAct functions cannot achieve. Therefore, it is important to select normalization methods that preserve the diversity of statistics across samples to maximize the advantages of LayerAct functions. With this consideration, batch-direction normalizations such as BN or Decorrelated Batch Normalization (DBN; \citet{Huang2018decorr}) are identified as the most effective choices for CNNs with LayerAct functions. Conversely, other normalization methods like LN or Switch Normalization (SwitchNorm; \citet{luo2019switchable}) are less favored. For the details and experiments, see Appendices G, H, and I. 

%%%%%%%%%%%%%%%%%%%%%%%%%%%%%%%%%%%%%%%%%%%%%%%%%%%%%%%%%%%%%%%%%
%%%%%%%%%%%%%%%%%%%%%%%%%%%%%%%%%%%%%%%%%%%%%%%%%%%%%%%%%%%%%%%%%
%%%%%%%%%%%%%%%%%%%%%%%%%%%%%%%%%%%%%%%%%%%%%%%%%%%%%%%%%%%%%%%%%

\section{Experiments}
\label{section_experiment}

In this section, we present the experimental analysis and results of LayerAct. For the detailed experimental setting to ensure reproducibility, see Appendix F.

%%%%%%%%%%%%%%%%%%%%%%%%%%%%%%%%%%%%%%%%%%%%%%%%%%%%%%%%%%%%%%%%%

\subsection{Experimental Analysis on MNIST}
\label{subsection_experimental_analysis}

In this subsection, we empirically analyze the properties of LayerAct discussed in the previous section: (1) diverse last layer representation between samples of different labels (2) activation robustness, and (3) zero-like activation mean. We trained a network with a single layer that containing 512 elements on the MNIST training dataset without any noise to observe the behavior of LayerAct functions during training. 

\subsubsection{Ensuring Diverse Representation}

\begin{figure}[t]
\centering
\includegraphics[width=0.49\columnwidth]{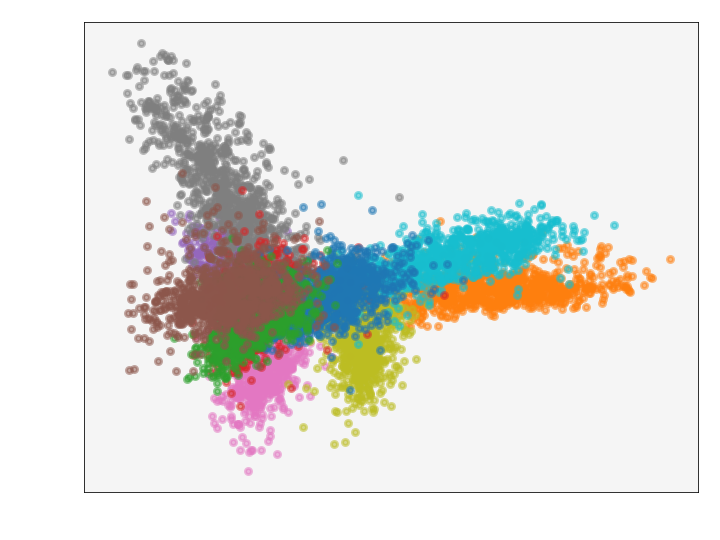}
\includegraphics[width=0.49\columnwidth]{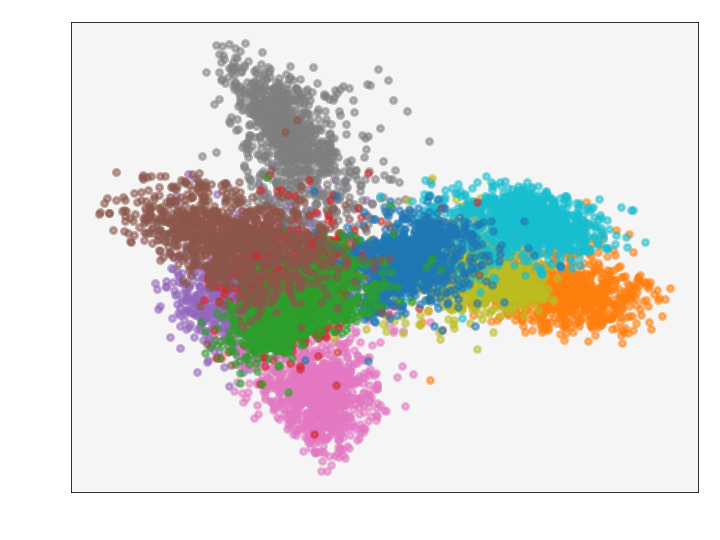}
\caption{Plots of the last layer representation in ResNet20 with LA-SiLU, with no normalization (left) and BN (right).}
\label{fig_iso_LayerAct}
\end{figure}

One of the key properties of LayerAct is that it does not produce homogeneous outputs, unlike LN, ensuring diverse representations in networks. Figure \ref{fig_iso_LayerAct} shows the final layer representation of ResNet20 with LA-SiLU without normalization or BN. LayerAct has more diverse representations than ResNet20 with LN and ReLU (Figure \ref{fig_iso_BN_LN}), despite utilizing layer-direction normalization. This is due to the activation output, which is the product of the activation input $y$ and the activation scale $s\left(n\right)$ (see Equation \ref{eq_layeract_forward}). This mechanism allows networks with BN and LayerAct to achieve better performance on clean datasets compared to those using LN.

\begin{figure}[t]
\centering
\includegraphics[width=1\columnwidth]{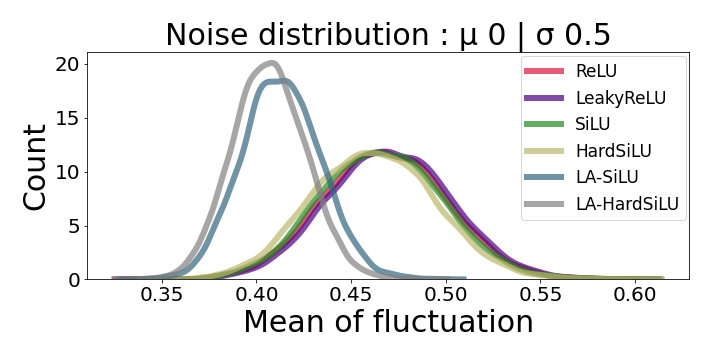} 
\caption{Distribution of activation output fluctuation.}
\label{fig_MNIST_fluctuation_output}
\end{figure}

\subsubsection{Activation Robustness}
\label{subsubsection_noise_robustness}

To confirm the robustness of LayerAct functions, we computed the activation fluctuation as defined in Definitions \ref{def_activation_fluctuation_element} and \ref{def_activation_fluctuation_layeract} using the network trained on the clean MNIST dataset. For the noisy input $\hat{y}_{i}$, we used two different noises with a normal distribution: one with a mean of zero and a variance of $0.5^{2}$, and another with a mean of one and a variance of $0.5^{2}$.

Figure \ref{fig_MNIST_fluctuation_output} shows the distribution of the activation fluctuation with a noise distribution. Although the fluctuation distribution of the activation input was similar (see Appendix K), LayerAct functions have a significantly smaller mean and variance of activation fluctuation among the samples than any other ElementAct function in all cases. The decrease in variance is remarkable, showing that LayerAct functions are robust for all samples. Moreover, the ElementAct functions that ensure a zero-like activation mean with one-sided saturation such as SiLU or HardSiLU showed slightly larger activation fluctuations than those of ReLU or LReLU when the noise had a large mean. However, LayerAct functions maintained lower fluctuations in both cases.

\begin{figure}[t]
\centering
\includegraphics[width=1\columnwidth]{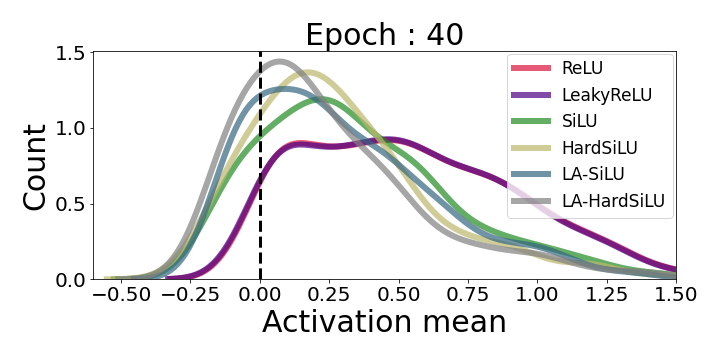} 
\caption{Distribution of the activation output means of the elements in a trained network on MNIST at epoch $40$.}
\label{fig_MNIST_activation_mean}
\end{figure}

\subsubsection{Zero-like Activation Mean}
\label{subsubsection_zero_like_mean}

Figure \ref{fig_MNIST_activation_mean} demonstrates the distribution of the activation output means of the single-layer network trained on the MNIST dataset at the epoch $40$. The distributions did not change after the epoch $40$. LayerAct functions maintain a zero-like activation mean for all epochs. Our experimental results indicate that LayerAct allows similar (before epoch $20$) or larger (after epoch $40$) negative outputs compared to other ElementAct functions. Thus, LA-SiLU and LA-HardSiLU can achieve a more zero-like activation mean than other activation functions.

%%%%%%%%%%%%%%%%%%%%%%%%%%%%%%%%%%%%%%%%%%%%%%%%%%%%%%%%%%%%%%%%%

\begin{table*}[t!]
\centering
\begin{tabular}{@{}llllllll@{}}
\hline
\hline
    & \textbf{CIFAR10} & \multicolumn{6}{c}{\bf CIFAR10-C} \\
Activation & Clean & Total & Noise & Blur & Digital & Weather & Extra \\
\hline
ReLU & 91.3 (0.045) & 68.8 (1.735) & 50.3 (5.967) & 65.2 (2.294) & 72.7 (0.948) & 78.6 (0.523) & 72.6 (1.471) \\
LReLU & 91.3 (0.069) & 68.7 (1.595) & 49.9 (5.175) & 65.3 (2.234) & 72.7 (0.801) & 78.4 (0.481) & 72.5 (1.464) \\
PReLU & 90.8 (0.054) & 67.9 (2.010) & 50.0 (6.180) & 64.0 (1.990) & 71.9 (1.700) & 77.3 (0.669) & 71.7 (1.787) \\
SiLU & \textbf{91.5} (0.042) & 69.0 (1.225) & 50.2 (3.131) & 65.6 (1.815) & 72.5 (0.763) & 78.9 (0.492) & 73.0 (1.015) \\
HardSiLU & 91.1 (0.052) & 68.5 (1.453) & 49.7 (5.605) & 65.2 (1.971) & 72.0 (0.652) & 78.2 (0.418) & 72.8 (1.191) \\
MISH & 91.5 (0.053) & 69.0 (1.324) & 50.0 (4.568) & 65.7 (1.720) & 72.6 (0.627) & 78.9 (0.401) & 72.9 (1.267) \\
GELU & \textbf{91.5} (0.035) & 68.6 (1.290) & 49.8 (3.318) & 64.9 (2.311) & 72.5 (0.729) & 78.7 (0.276) & 72.6 (1.276) \\
ELU & 91.0 (0.030) & 68.7 (1.363) & 48.3 (5.193) & 66.5 (1.842) & 72.3 (0.605) & 78.8 (0.355) & 72.5 (1.220) \\
\hline
LA-SiLU & \textbf{\underline{91.6}} (0.027) & \textbf{\underline{70.4}} (1.392) & \textbf{51.8} (3.527) & \textbf{\underline{67.5}} (2.155) & \textbf{\underline{74.0}} (0.948) & \textbf{\underline{79.9}} (0.559) & \textbf{\underline{74.4}} (1.014) \\
LA-HardSiLU & 91.2 (0.042) & \textbf{70.3} (1.384) & \textbf{\underline{52.2}} (5.707) & \textbf{67.5} (1.238) & \textbf{73.5} (0.836) & \textbf{79.5} (0.501) & \textbf{74.3} (1.127) \\
\hline
\hline
\end{tabular}
\caption{Classification performance of ResNet20 on the CIFAR10 and CIFAR10-C. We present mean and (variance).}
\label{table_experiment_cifar10_resnet20}
\end{table*}

\begin{table*}[tb]
\centering
\begin{tabular}{@{}lllllllll@{}}
\hline
\hline
    & & \textbf{ImageNet} & \multicolumn{6}{c}{\bf ImageNet-C} \\
    Model & Activation & Clean & Total & Noise & Blur & Digital & Weather & Extra  \\ 
\hline
    ResNet50 & ReLU & 77.71 & \textbf{43.75} & \textbf{34.40} & 36.85 & \textbf{48.37} & 47.18 & 49.60 \\
    ResNet50 & LReLU & 77.83 & 43.24 & 32.87 & 36.33 & 48.00 & 47.03 & 49.38 \\
    ResNet50 & PReLU & 74.99 & 36.77 & 23.28 & 32.27 & 43.29 & 39.56 & 42.05 \\
    ResNet50 & SiLU & 77.85 & 42.31 & 29.74 & 35.94 & 46.59 & 47.36 & 48.77 \\
    ResNet50 & HardSiLU & 76.30 & 40.56 & 26.21 & 35.14 & 46.01 & 45.23 & 46.63 \\
    ResNet50 & Mish & 77.41 & 42.57 & 31.24 & 36.60 & 46.73 & 46.83 & 48.64 \\
    ResNet50 & GELU & 78.01 & 40.71 & 27.82 & 35.35 & 45.34 & 44.52 & 47.28 \\
\hline
    ResNet50 & LA-SiLU & \textbf{\underline{78.62}} & \textbf{\underline{45.29}} & \textbf{\underline{36.16}} & \textbf{\underline{37.66}} & \textbf{\underline{50.31}} & \textbf{\underline{48.33}} & \textbf{\underline{51.71}} \\
    ResNet50 & LA-HardSiLU & \textbf{78.24} & 43.63 & 32.21 & \textbf{37.57} & 47.69 & \textbf{47.88} & \textbf{49.93} \\
\hline
\hline
    ResNet101 & ReLU & \textbf{\underline{79.24}} & 46.7 & 36.29 & 39.75 & 52.16 & 49.93 & 52.78 \\
    ResNet101 & LA-SiLU & 79.12 & \textbf{\underline{47.87}} & \textbf{\underline{39.62}} & \textbf{\underline{39.85}} & \textbf{\underline{52.82}} & \textbf{\underline{50.84}} & \textbf{\underline{54.16}} \\
\hline
\hline
\end{tabular}
\caption{Classification performance of ResNet50 and ResNet101 on the clean and noisy ImageNet.}
\label{table_experiment_ImageNet}
\end{table*}

\subsection{Classification Performance}
\label{subsection_classification_performance}

We demonstrate the classification performance of LayerAct on CIFAR10, ImageNet \cite{russakovsky2015imagenet}, and a medical image dataset. We used ResNet \cite{he2016deep} as the network for our experiments on the three benchmark datasets. In our experiments on CIFAR10 and ImageNet, we utilized BN. We compared LayerAct with ReLU, LReLU, PReLU, SiLU, HardSiLU, Mish \cite{misra2020mish}, GELU \cite{hendrycks2016gaussian}, and ELU \cite{clevert2015fast}. In the tables, The best results are underlined and bolded, and the second best are bolded. In the main manuscript, we present the results of ResNet20 on CIFAR10 and ResNet50 and ResNet101 on ImageNet. For more experimental results, including on CIFAR100, see Appendix J.

\subsubsection{On CIFAR10 and CIFAR10-C} 

The column named clean of Table \ref{table_experiment_cifar10_resnet20} presents the classification performance of ResNet20 with both LayerAct functions and ElementAct functions, benchmarked on the clean CIFAR10. We report the mean accuracy over 30 runs. LA-SiLU achieved the best performance among the activation functions. In other experiments, with ResNet32 and ResNet44 on clean CIFAR10 (Appendix J), networks with GELU achieved better results in two cases: ResNet32 on CIFAR10 and ResNet44 on CIFAR100. In the remaining combinations of networks and datasets, LA-SiLU outperformed in a significant majority of cases, especially in 39 out of 48 cases according to the $p$-value from a T-test or a Wilcoxon signed-rank test, indicating statistical significance.

To verify the robustness of LayerAct functions, we evaluated their classification performance on CIFAR10-C. The columns other except clean in Table \ref{table_experiment_cifar10_resnet20} demonstrates the experimental results. The networks with LayerAct functions achieved remarkable performance compared to those with ElementAct functions. Specifically, LA-SiLU and LA-HardSiLU statistically outperformed other ElementAct functions in 46 out of 48 cases for each, according to the $p$-value from a T-test or a Wilcoxon signed-rank test. This shows that LayerAct functions exhibit greater robustness to intense noise compared to other functions. Furthermore, the experimental results on ResNets without a normalization method reveal that LA-SiLU can maintain its robustness when inputs are not excessively large. 

\subsubsection{On ImageNet} 

Table \ref{table_experiment_ImageNet} shows the performance of ResNet50 and ResNet101 with LayerAct functions and other ElementAct functions for comparison with clean and noisy ImageNet datasets. For the column named "clean," we report the accuracy of 10-crop testing on the validation dataset. We used the out-of-distribution benchmark dataset, ImageNet-C \cite{hendrycks2019robustness}, which has the same corruptions as CIFAR-C datasets. The networks with LayerAct functions outperformed those with other activation functions on all datasets. The ResNet50 with LayerAct functions, even with LA-HardSiLU which showed worse performance on the clean CIFAR10 datasets compared to SiLU or LReLU, outperformed other activation functions on clean ImageNet. Based on these results, we trained deeper networks, ResNet101, utilizing ReLU and LA-SiLU as activations. LA-SiLU maintained its superior robustness.

\begin{table}[tb]
\centering
\begin{tabular}{@{}llll@{}}
\hline
\hline
    Activation & U-net & Unet++ w/o DSV & Unet++ w DSV\\
\hline
    ReLU & 84.71 & 84.94 & 84.92 \\
    SiLU & 84.87 & 85.15 & 85.01 \\
    LA-SiLU & \textbf{85.13} & \textbf{85.27} & \textbf{85.05} \\
\hline
\hline
\end{tabular}
\begin{tabular}{@{}lllllll@{}}
Model & \multicolumn{2}{c}{ReLU} & \multicolumn{2}{c}{SiLU} & \multicolumn{2}{c}{LA-SiLU} \\ 
& Nat & Adv & Nat & Adv & Nat & Adv \\
\hline
RN18 & 81.79 & 82.73 & 83.28 & 82.93 & \textbf{83.64} & \textbf{84.14} \\
\hline
\hline
\end{tabular}
\caption{Additional experiments on medical image segmentation (upper) and adversarial robustness (lower).}
\label{table_additional_ex}
\end{table}

\subsubsection{Additional Experiments} 

To evaluate LayerAct's effectiveness, we conducted additional experiments on a nuclei image dataset \cite{data-science-bowl-2018} for segmentation and on CIFAR10 to verify adversarial robustness. Table \ref{table_additional_ex} shows that networks with LA-SiLU outperform those with ReLU or SiLU in both tasks. Results from segmentation tasks with U-Net \cite{ronneberger2015u} and UNet++ \cite{zhou2018unet++} show LayerAct's potential across different architectures for image tasks. The experimental results on adversarial robustness with ResNet18 (RN18) further underscore LayerAct's robustness. For the detail, see Appendices L and M for experiments on medical image and adversarial robustness, respectively. 

\section{Conclusion}

In this work, we identified the fundamental limitation of ElementAct in robustness and demonstrated that layer-direction normalization has the potential to enhance the robustness of activation. Inspired by our investigation, we have developed LayerAct mechanism and functions that not only enhance robustness compared to existing activation functions but also resolve the trade-off problem. Moreover, unlike LN, LayerAct does not reduce the performance of CNNs when utilized with BN on clean dataset. Experimental results on benchmark datasets show that LayerAct functions preserve essential activation properties and provide robust performance on both clean and noisy datasets, underscoring their utility and effectiveness in advanced CNN architectures.

%%%%%%%%%%%%%%%%%%%%%%%%%%%%%%%%%%%%%%%%%%%%%%%%%%%%%%%%%%%%%%%%%
%%%%%%%%%%%%%%%%%%%%%%%%%%%%%%%%%%%%%%%%%%%%%%%%%%%%%%%%%%%%%%%%%
%%%%%%%%%%%%%%%%%%%%%%%%%%%%%%%%%%%%%%%%%%%%%%%%%%%%%%%%%%%%%%%%%

\section*{Acknowledgements}

This research was partly supported by the Basic Science Research Program through the National Research Foundation of Korea (NRF), funded by the Ministry of Education (RS-2024-00463188) and the Ministry of Science and ICT (RS-2024-00458720), as well as by the Institute of Information \& Communications Technology Planning \& Evaluation (IITP) grants funded by the Korean government (MSIT) (No.RS-2024-00439932, SW Starlab; No.RS-2020-II201336, Artificial Intelligence Graduate School Program - UNIST; No.RS-2021-II212068, Artificial Intelligence Innovation Hub; No.RS-2024-00443780, Development of Foundation Models for Bioelectrical Signal Data and Validation of Their Clinical Applications: A Noise-and-Variability Robust, Generalizable Self-Supervised Learning Approach).

%%%%%%%%%%%%%%%%%%%%%%%%%%%%%%%%%%%%%%%%%%%%%%%%%%%%%%%%%%%%%%%%%
%%%%%%%%%%%%%%%%%%%%%%%%%%%%%%%%%%%%%%%%%%%%%%%%%%%%%%%%%%%%%%%%%
%%%%%%%%%%%%%%%%%%%%%%%%%%%%%%%%%%%%%%%%%%%%%%%%%%%%%%%%%%%%%%%%%

\bibliography{aaai25}

%%%%%%%%%%%%%%%%%%%%%%%%%%%%%%%%%%%%%%%%%%%%%%%%%%%%%%%%%%%%%%%%%
%%%%%%%%%%%%%%%%%%%%%%%%%%%%%%%%%%%%%%%%%%%%%%%%%%%%%%%%%%%%%%%%%
%%%%%%%%%%%%%%%%%%%%%%%%%%%%%%%%%%%%%%%%%%%%%%%%%%%%%%%%%%%%%%%%%

\clearpage
\appendix

\newgeometry{left=30mm, right=30mm, top=30mm, bottom=40mm}

\input{LayerAct_appendix}

\restoregeometry

\end{document}

%% file: LayerAct_appendix.tex
\title{Appendix} 
\author{LayerAct: Advanced Activation Mechanism for Robust Inference of CNNs}
\date{}

\maketitle

\appendix
\allowdisplaybreaks

\section{A. Important Properties of Activation}

One-side saturation and zero-like activation mean stand out as crucial properties of activation for effective and efficient network training. Activation functions that exhibit one-side saturation, such as the ReLU, are favored within deep networks for their potential to provide more informative signal propagation during the backward pass by allowing for larger derivatives \cite{glorot2011deep}. Furthermore, the presence of a saturation state contributes to robustness against input shifts, maintaining consistent activation outputs for elements in this state and thereby stabilizing network training \cite{clevert2015fast, qiu2018frelu}.

Achieving a zero-like activation mean is another important property of activation functions, facilitating efficient network training. The activation mean refers to the average activation output of an individual element across samples. Functions such as the Exponential Linear Unit (ELU) \cite{clevert2015fast} and Parametric ReLU (PReLU) \cite{he2015delving} push the activation mean toward zero by allowing negative outputs. Activation functions with this property can correct the bias shift that occurs between layers and lead to efficient network training \cite{clevert2015fast, qiu2018frelu}.

The activation output of the $i^{th}$ unit of $m^{th}$ sample ($m\in\{1,2,...,M\}$) is defined as $a_{m,i}=f\left(y_{m,i}\right)$, where $f$, $y_{m,i}$, and $M$ are activation function, the $i^{th}$ activation input of the $m^{th}$ sample, and the number of samples in the batch, respectively. Ideally, a ``zero-like activation mean'' occurs when the activation mean of a single unit, $a_{i}$, approximates zero across the samples. Mathematically, this can be represented as:
\begin{equation*}
\label{eq_zero_like_activaiton_mean}
    \begin{split}
        \frac{1}{M}\sum^{M}_{m=1} a_{m,i} \approx 0
    \end{split}.
\end{equation*}
However, approximating the activation mean to zero is challenging for activation functions that saturate at (large) negative outputs, such as ELU, SiLU, or FReLU. Due to this saturation, previous studies have defined the ``zero-like activation mean'' property of an activation function as its ability to ``push'' the activation mean towards zero. In mathematical terms, this can be presented as $\left\vert \mu_{a_{i}}\right\vert \ll c$, where $c$ is a small positive constant \cite{clevert2015fast, qiu2018frelu}.

%%%%%%%%%%%%%%%%%%%%%%%%%%%%%%%%%%%%%%%%%%%%%%%%%%%%%%%%%%%%%%%%%%%%%%

\section{B. Detailed Derivation Process of Activation Fluctuation with LN}

In this section, we present the detailed derivation process of the equations in Definition 4 in the main manuscript as follows:
\begin{align*}
    \left\Vert f\left(\hat{n}\right) - f\left(n\right) \right\Vert
    =& 
    \sum^{D}_{i=1}
    \left\vert 
        \hat{n}_{i} s\left(\hat{n}_{i}\right)
        - n_{i}s\left(n_{i}\right)
    \right\vert 
    \\
    =&
    \sum^{D}_{i=1}
    \left\vert
        \frac{\hat{y}_{i} + \epsilon_{i} -\mu_{y} - \mu_{\epsilon}}{\sqrt{\sigma^{2}_{y} + \sigma^{2}_{\epsilon} + \alpha}} s\left(\hat{n}_{i}\right)
        -
        \frac{y_{i}-\mu_{y}}{\sqrt{\sigma^{2}_{y}+\alpha}} s\left(n_{i}\right)
    \right\vert
    \\
    \approx&
    \sum^{D}_{i=1}
    \left\vert
        n_{i}\left(s\left(\hat{n}_{i}\right)-s\left(n_{i}\right)\right)
        +
        \frac{\left(\epsilon - \mu_{\epsilon}\right)s\left(\hat{n}_{i}\right)}{\sqrt{\sigma_{y}^{2}+\alpha}}
    \right\vert 
    \\
    \leq&
    \sum^{D}_{i=1}
    \left(
        \left\vert n_{i} \right\vert \cdot
        \left\vert
            s\left(\hat{n}_{i}\right)-s\left(n_{i}\right)
        \right\vert
        +
        \frac{\left\vert\epsilon - \mu_{\epsilon}\right\vert \cdot s\left(\hat{n}_{i}\right)}{\sqrt{\sigma_{y}^{2}+\alpha}}
    \right),  
\end{align*}
where $\sqrt{\sigma_{y}^{2}+\alpha+\sigma_{\epsilon}^{2}} \approx \sqrt{\sigma_{y}^{2}+\alpha}>1$ when $\sigma_{y}\gg\sigma_{\epsilon}$ and $\alpha$ is sufficiently large. 

%%%%%%%%%%%%%%%%%%%%%%%%%%%%%%%%%%%%%%%%%%%%%%%%%%%%%%%%%%%%%%%%%%%%%%

\section{C. Difference between LayerAct and Activation with LN}

In this section, we provide detailed comparison between the activation outputs of LayerAct and those of element-level activation functions paired with LN. When a LN layer is placed right before an activation layer, the $i^{th}$ activation output is $a_{i}=n^{LN}_{i}s\left(n^{LN}_{i}\right)$, where $n^{LN}_{i}$ is the $i^{th}$ normalized output of LN. Conversely, the $i^{th}$ activation output of a LayerAct function is $a_{i}=y_{i}s\left(n_{i}\right)$, as defined in Equation 12 in the main manuscript. 

The critical distinction between activation with LN and LayerAct lies in the preservation of input mean and variance statistical information in the activation output. \textit{LayerAct can preserve the statistical information, but activation with LN cannot.} With LN, the activation function takes a normalized input in layer-direction, resulting in activation outputs that exhibit similar mean and variance across samples (as shown in the activation output equation for LN above). This homogenization of statistical information across samples, a characteristic of LN, is a reason why BN often outperforms LN in non-sequential models such as CNNs \cite{labatie2021proxy, Lubana2021Beyond}. In the main manuscript, we demonstrated that this homogeneity limitation of LN leads the representation of the samples with different labels to be similar (see Figure 1 in the main manuscript). 

LayerAct, on the other hand, produces more distinguishable activation outputs between samples by preserving statistical variation between samples. This is due to the fact that only the activation scale function of LayerAct uses the layer-normalized input, not the LayerAct function itself (as shown in Equation 12 in the main manuscript). Based on this unique output of LayerAct, networks with LayerAct can produce much more distinguishable representation of samples compared to those with activation and LN (see Figure 3 in the main manuscript). 

%%%%%%%%%%%%%%%%%%%%%%%%%%%%%%%%%%%%%%%%%%%%%%%%%%%%%%%%%%%%%%%%%%%%%%

\begin{figure}[ht]
\centering
\includegraphics[width=0.22\columnwidth]{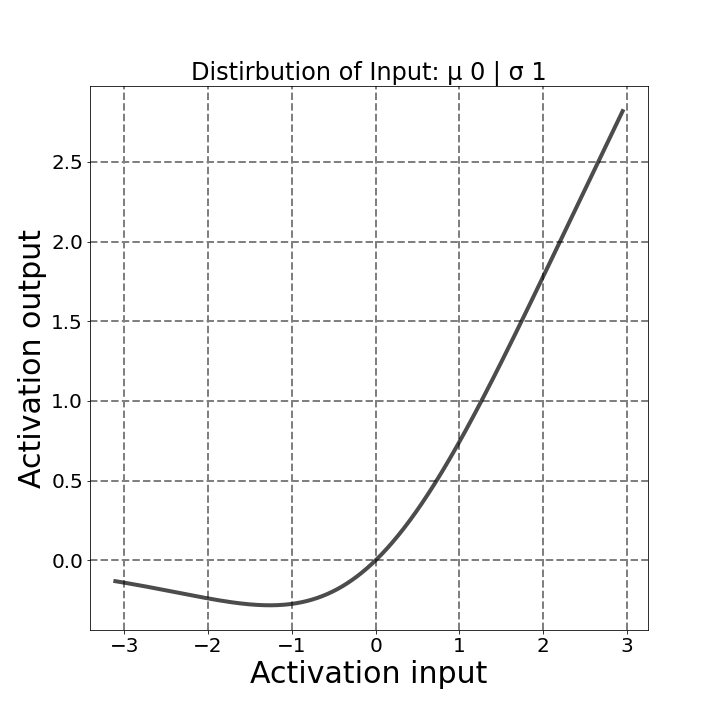} 
\includegraphics[width=0.22\columnwidth]{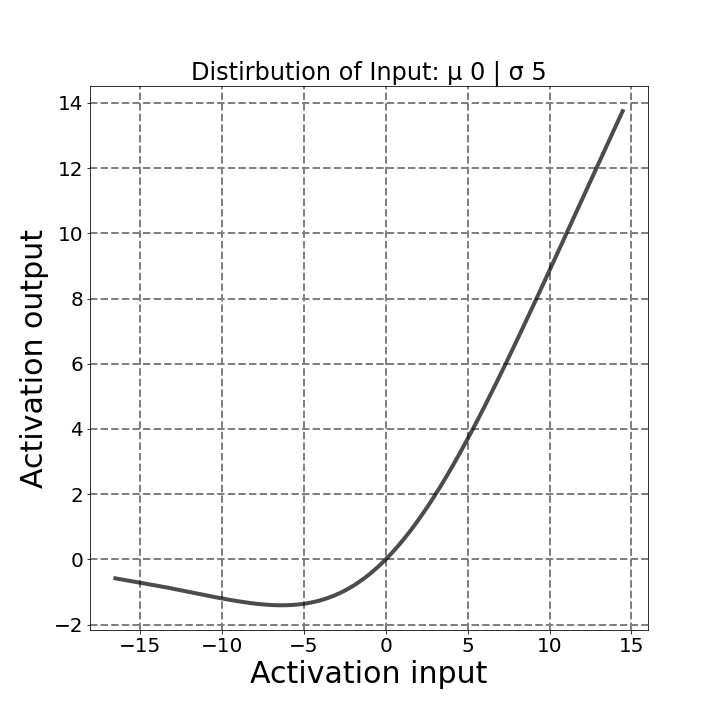} 
\includegraphics[width=0.22\columnwidth]{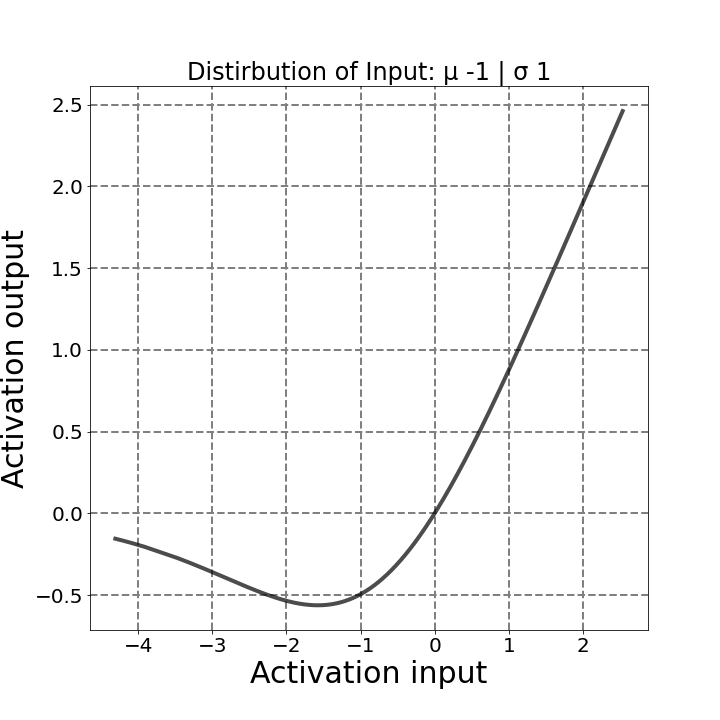} 
\includegraphics[width=0.22\columnwidth]{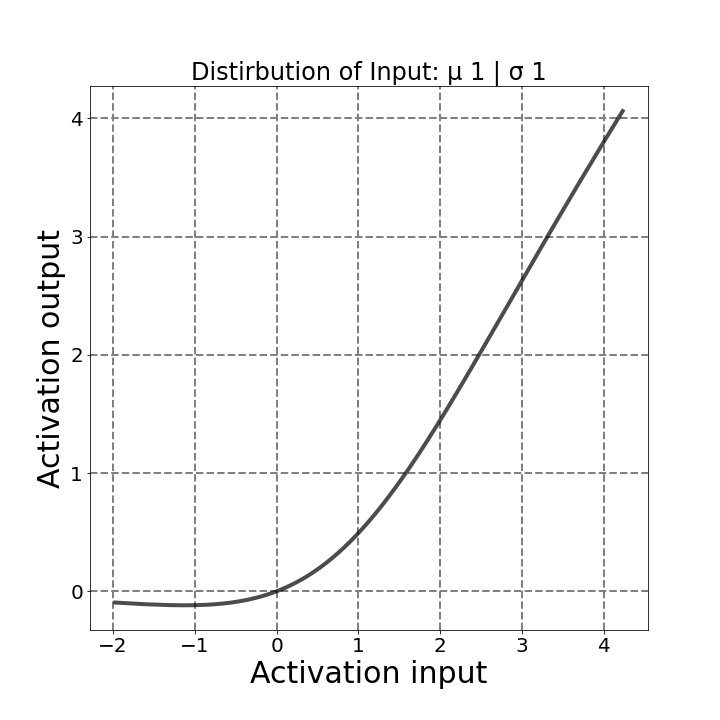} 
\caption{LA-SiLU with different mean and variance values in the input. The distribution of the activation input is: (1) $\mu_{y}=0$, $\sigma_{y}=1$, (2) $\mu_{y}=0$, $\sigma_{y}=5$, (3) $\mu_{y}=-5$, $\sigma_{y}=1$, and (4) $\mu_{y}=5$, $\sigma_{y}=1$ from the left to right.}
\label{fig_LA-SiLU_activation_output}
\end{figure}

\begin{figure}[ht]
\centering
\includegraphics[width=0.22\columnwidth]{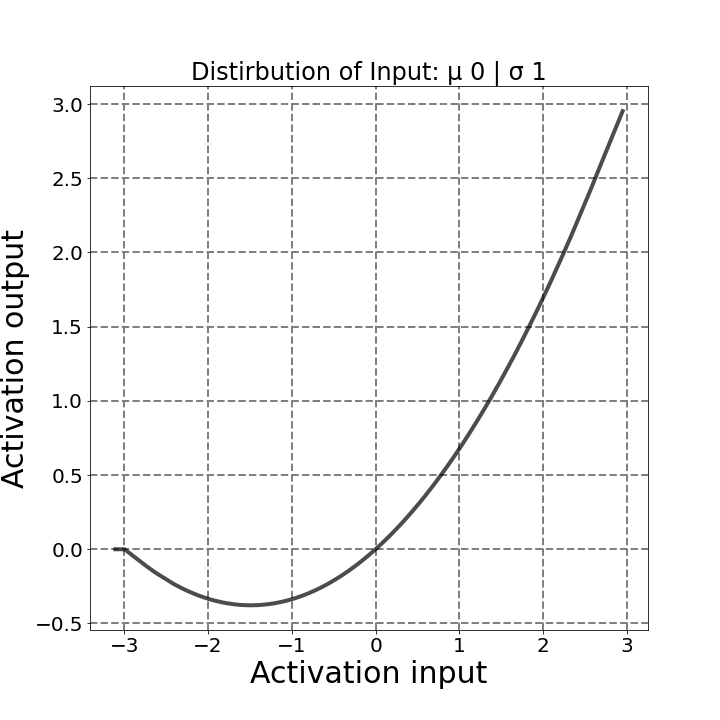} 
\includegraphics[width=0.22\columnwidth]{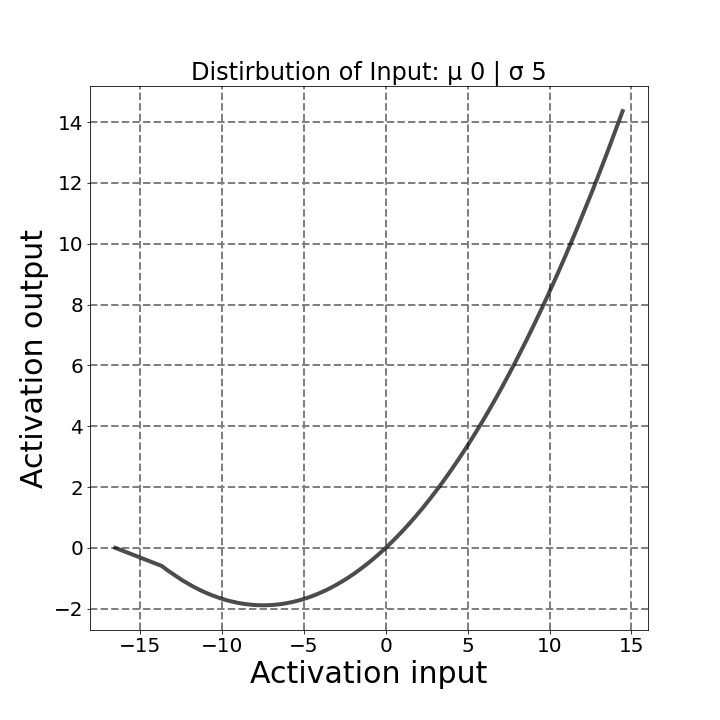} 
\includegraphics[width=0.22\columnwidth]{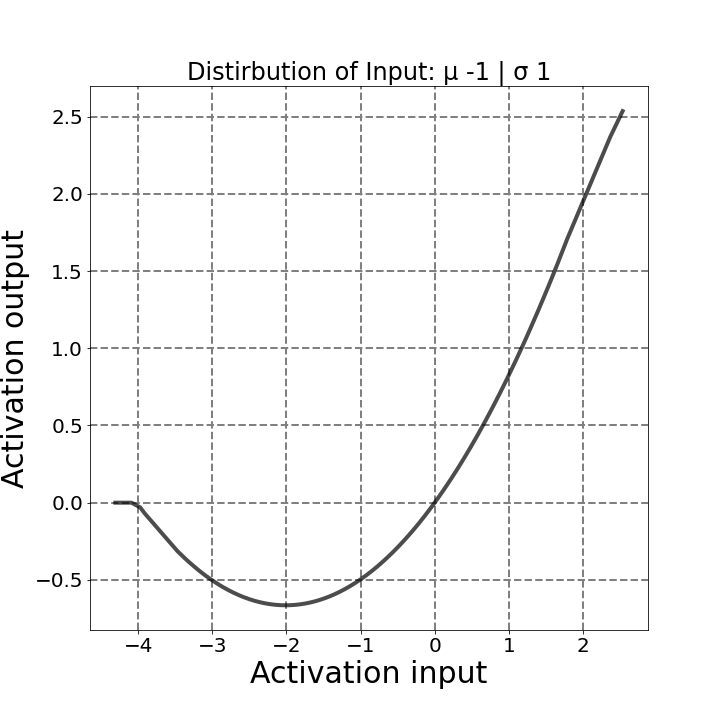} 
\includegraphics[width=0.22\columnwidth]{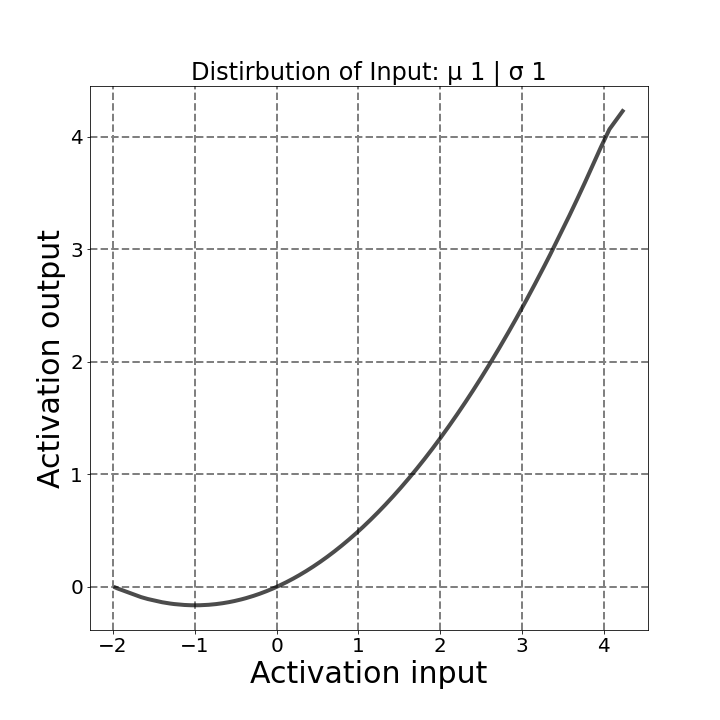} 
\caption{LA-HardSiLU with different mean and variance values in the input. The distribution of the activation input is: (1) $\mu_{y}=0$, $\sigma_{y}=1$, (2) $\mu_{y}=0$, $\sigma_{y}=5$, (3) $\mu_{y}=-5$, $\sigma_{y}=1$, and (4) $\mu_{y}=5$, $\sigma_{y}=1$ from the left to right.}
\label{fig_LA-HardSiLU_activation_output}
\end{figure}

\section{D. Activation Output of LayerAct Functions}

In this section, we provide detailed comparison between the activation outputs of LayerAct and those of element-level activation functions paired with LN. When a LN layer is placed right before an activation layer, the $i^{th}$ activation output is $a_{i}=n^{LN}_{i}s\left(n^{LN}_{i}\right)$, where $n^{LN}_{i}$ is the $i^{th}$ normalized output of LN. Conversely, the $i^{th}$ activation output of a LayerAct function is $a_{i}=y_{i}s\left(n_{i}\right)$, as defined in Equation 12 in the main manuscript. 

The critical distinction between activation with LN and LayerAct lies in the preservation of input mean and variance statistical information in the activation output. \textit{LayerAct can preserve the statistical information, but activation with LN cannot.} With LN, the activation function takes a normalized input in layer-direction, resulting in activation outputs that exhibit similar mean and variance across samples (as shown in the activation output equation for LN above). This homogenization of statistical information across samples, a characteristic of LN, is a reason why BN often outperforms LN in non-sequential models such as CNNs \cite{labatie2021proxy, Lubana2021Beyond}. 

LayerAct, on the other hand, produces more distinguishable activation outputs between samples by preserving statistical variation between samples. This is due to the fact that only the activation scale function of LayerAct uses the layer-normalized input, not the LayerAct function itself (as shown in Equation 12 in the main manuscript). 

%%%%%%%%%%%%%%%%%%%%%%%%%%%%%%%%%%%%%%%%%%%%%%%%%%%%%%%%%%%%%%%%%%%%%%

\section{E. Trade-off between Two Properties}

A fundamental trade-off exists between the important properties of activation, one-side saturation and zero-like mean activation. Activation functions that have a saturation state typically limit negative outputs, consequently driving the activation mean away from zero. On the other hand, to achieve a zero-like activation mean, certain functions like PReLU allow large negative outputs. However, functions with this allowance are not robust due to the absence of saturation.

To address this trade-off issue, various activation functions, such as Sigmoid Linear Unit (SiLU, also known as SWISH) \cite{elfwing2018sigmoid, ramachandran2017searching}, Gaussian Error Linear Unit (GELU) \cite{hendrycks2016gaussian}, and Mish \cite{misra2020mish}, saturate the large negative outputs only. These activation functions can achieve a zero-like activation mean with small negative outputs. However, a trade-off still exists because the restriction of negative outputs, designed to ensure saturation, prevents the allowance of large negative outputs, thereby restraining the mean of activation from being more zero-like.

\textit{Additionally, this trade-off cannot be addressed by LN}. Even when a LN layer is placed before the activation layer, the large negative output is restricted with element-level activation. Therefore, mitigating the trade-off between activation properties is a distinct benefit of LayerAct functions that element-level activation functions cannot attain.

\section{F. Experimental Setting}
\label{app_sec_experimental_setting}

We implemented LayerAct functions and networks for experiment with PyTorch \cite{paszke2019pytorch}. All networks used in our experiments were trained on NVIDIA A100. We used multiple devices to train the networks on ImageNet, and a single device for the other experiments. The versions of Python and the packages were (1) Python 3.9.12, (2) numpy 1.19.5 (3) PyTorch 1.11.0, and (4) torchvision 0.12.0. We used cross entropy loss functions for all the experiments. The random seeds of the experiments were $11\times i$ where $i\in\{1,2,...,30\}$ on CIFAR10 and CIFAR100 and $11$ on ImageNet.

To train networks on MNIST for experimental analysis, we applied batch gradient descent for $80$ epochs with the weight decay and momentum fixed to $0.0001$ and $0.9$, respectively. The learning rate started from 0.01, and was multiplied $0.1$ at epochs $40$ and $60$ as scheduled. We used ResNet \cite{he2016deep} with BN right before activation for experiments on CIFAR10, CIFAR100 and ImageNet. We initialized the weights following the methods proposed by He et al. \cite{he2015delving}. For all experiments, the weight decay, momentum, and initial learning rate were $0.0001$, $0.9$ and $0.1$, respectively.

For CIFAR10 and CIFAR100, we trained ResNet20, ResNet32, and ResNet44 with a basic block using the stochastic gradient descent with a batch size of 128 for about 64000 iterations. We randomly selected $10\%$ of the training dataset as the validation set. The learning rate was scheduled to decrease by the factor of 10 at 32000 and 48000 iterations. For the data augmentation of CIFAR10 and CIFAR100, we followed \cite{lee2015deeply}. We rescaled the data between $0$ and $1$, padded $4$ pixels on each side, and randomly sampled a $32\times32$ crop from the padded image or its horizontal flip. The data was normalized after augmentation. For testing, we did not apply data augmentation, only normalized the data. The hyper-parameter $\alpha$ of LayerAct functions for the experiments was set to $0.00001$.

For the experiment with ImageNet, we trained ResNet50 and ResNet101 with the bottleneck block using stochastic gradient descent, and the batch size was 256 for about 600000 iterations. The learning rate was scheduled to decrease by a factor of 10 at 180000, 360000, and 540000 iterations. For the data augmentation on ImageNet, we rescaled the data between $0$ and $1$, resized it to $224\times244$, and randomly sampled a $224\times224$ crop from an image or its horizontal flip \cite{krizhevsky2017ImageNet}. We normalized the data after data augmentation. For testing, we resized the data to be $256\times256$ and applied 10-crop. Afterward, the data was normalized. To ensure stable learning, we set the hyper-parameter $\alpha$ of LayerAct functions to $0.1$ which is larger than those for CIFAR10 and CIFAR100. 

For the reproducibility, we have attached the code that used for the experiments in supplementary materials. 

%%%%%%%%%%%%%%%%%%%%%%%%%%%%%%%%%%%%%%%%%%%%%%%%%%%%%%%%%%%%%%%%%%%%%%

\section{G. Relationship between LayerAct and BN}

In our main manuscript, we emphasized the importance of selecting an appropriate normalization method for employing LayerAct functions effectively. This section delves into the relationship between LayerAct and BN, the most dominant normalization method for CNNs. In this section, our objective is to discuss the benefits of BN in CNNs and demonstrate how LayerAct functions collaborate well with BN, in contrast to LN. 

BN operates along three dimensions, batch, height, and width, particularly for image datasets, where the input matrix $y\in\mathbb{R}^{B\times C \times H \times W}$. To simplify, we consider a flattened dimension that combines height and width, denoted by size $D$. The mean, variance, and output of BN are then defined as follows: 
\begin{align}
    \mu^{BN}_{j} =& \frac{1}{BD}\sum^{B}_{i=1}\sum^{D}_{k=1}y_{i,j,k}, 
    \\
    \sigma^{BN}_{j} =& \frac{1}{BD}\sum^{B}_{i=1}\sum^{D}_{k=1}\left(y_{i,j,k}-\mu^{BN}_{j}\right), 
    \\
    n^{BN}_{i,j,k} =& \gamma_{j} \cdot \frac{y_{i,j,k}-\mu^{BN}_{j}}{\sqrt{{\sigma^{BN}_{j}}^{2}+\alpha}} + \beta_{j}.
\end{align}

BN enjoys the common advantages of normalization methods, re-scaling and re-centering operations that significantly enhance the efficiency and effectiveness of network training, promote stable training, and allow the use of a larger learning rate \cite{Lubana2021Beyond, labatie2021proxy, ioffe2015Batch}.

Additionally, BN offers the unique benefit of avoiding channel collapse, a phenomenon where a channel exhibits a linear activation (i.e. a channel loses its non-linear activation), leading to inefficient network training \cite{labatie2021proxy}. For instance, when a channel's activation inputs are consistently positive or negative across all samples, the activation output of a ReLU layer becomes linear, $a=y$ or $a=0$, respectively. BN addresses this by normalizing across the batch dimension, ensuring a diverse distribution of channel activations among samples. Moreover, it does not have the homogenization issue present in LN.

When utilized with LayerAct, BN does not compromise the benefits of LayerAct because they operate on different normalization dimensions. Given that the output of LayerAct functions is the product of the activation input (i.e. the normalization output) and the activation scale, BN's benefits extend to the activation output, enhancing overall network performance as when utilized with element-level activation function. Therefore, utilizing LayerAct with BN preserves the advantages of both, enhancing the efficiency of network training and robustness of network inference.

In summary, the choice of normalization method for CNNs with LayerAct functions should be made carefully, taking into account the interplay between LayerAct and the normalization. While LN may diminish LayerAct's benefits, as discussed in Appendix \ref{app_layeract_with_LN}, BN, with its beneficial properties, would be a suitable choice for CNNs with LayerAct functions on image datasets. 

%%%%%%%%%%%%%%%%%%%%%%%%%%%%%%%%%%%%%%%%%%%%%%%%%%%%%%%%%%%%%%%%%%%%%%

\begin{table}[tb]
\caption{Classification performance on CIFAR and CIFAR-C datasets without a normalization method. C10 and C100 denotes CIFAR10 and CIFAR100, respectively.}
\label{table_appendix_without_norm}
\centering
\begin{tabular}{@{}llllllllll@{}}
\hline \hline
    & & & \textbf{CIFAR} & \multicolumn{6}{c}{\bf CIFAR-C} \\
    Data & Model & Activation & Clean & Total & Noise & Blur & Digital & Weather & Extra  \\ 
    \hline 
    C10 & ResNet20 & ReLU & 88.51 & 67.43 & 53.27 & 62.12 & 70.93 & 75.91 & 71.38 \\
    C10 & ResNet20 & SiLU & 88.29 & 68.94 & 54.66 & \textbf{65.84} & 71.6 & 76.54 & 72.48 \\
    C10 & ResNet20 & LA-SiLU & \textbf{88.95} & \textbf{69.36} & \textbf{54.71} & \textbf{65.58} & \textbf{71.93} & \textbf{77.59} & \textbf{73.34} \\
    \hline 
    C10 & ResNet32 & ReLU & \textbf{89.56} & 68.94 & 54.48 & 63.90 & 72.45 & 77.52 & 72.73 \\
    C10 & ResNet32 & SiLU & 46.51 & 38.69 & 33.21 & 37.52 & 39.62 & 41.6 & 40.11 \\
    C10 & ResNet32 & LA-SiLU & 89.12 & \textbf{70.59} & \textbf{56.29} & \textbf{66.81} & \textbf{73.11} & \textbf{78.94} & \textbf{74.22} \\
    \hline 
    C10 & ResNet44 & ReLU & \textbf{90.03} & 69.97 & 55.75 & 65.08 & 73.38 & 78.45 & 73.62 \\
    C10 & ResNet44 & SiLU & 15.25 & 14.18 & 13.5 & 14.01 & 14.26 & 14.55 & 14.39 \\
    C10 & ResNet44 & LA-SiLU & 88.77 & \textbf{71.65} & \textbf{58.15} & \textbf{68.32} & \textbf{74.01} & \textbf{79.49} & \textbf{74.88} \\
    \hline 
    C100 & ResNet20 & ReLU &59.46 &37.69 &21.17 &38.76 &41.36 &44.15 &38.89 \\
    C100 & ResNet20 & SiLU &59.74 &40.39 &24.21 &42.72 &43.51 &46.45 &41.01 \\
    C100 & ResNet20 & LA-SiLU &\textbf{61.01} &\textbf{42.44} &\textbf{26.07}  &\textbf{44.50} &\textbf{45.74} &\textbf{48.8} &\textbf{42.98} \\
    \hline 
    C100 & ResNet32 & ReLU &\textbf{60.71} &39.56 &23.05 &40.78 &43.23 &45.93 &40.67 \\
    C100 & ResNet32 & SiLU &20.36 &14.99 &10.05 &16.27 &15.89 &16.47 &15.02 \\
    C100 & ResNet32 & LA-SiLU &60.08 &\textbf{44.01} &\textbf{28.77} &\textbf{46.73} &\textbf{47.02} &\textbf{49.51} &\textbf{44.19} \\
    \hline 
    C100 & ResNet44 & ReLU &\textbf{61.59} &40.88 &24.57 &42.08 &44.60 &47.16 &41.90 \\
    C100 & ResNet44 & SiLU & 2.91 & 2.43 & 1.97 & 2.58 & 2.53 & 2.54 & 2.42 \\
    C100 & ResNet44 & LA-SiLU &60.30 &\textbf{44.26} &\textbf{29.22} &\textbf{46.74} &\textbf{47.45} &\textbf{49.65} &\textbf{44.50} \\
\hline \hline
\end{tabular}
\end{table}

\begin{table}[tb]
\caption{Classification performance on CIFAR and CIFAR-C datasets with LN. C10 and C100 denote CIFAR10 and CIFAR100, respectively.}
\label{table_appendix_with_LN}
\centering
\begin{tabular}{@{}llllllllll@{}}
\hline \hline
    & & & \textbf{CIFAR} & \multicolumn{6}{c}{\bf CIFAR-C} \\
    Data & Model & Activation & Clean & Total & Noise & Blur & Digital & Weather & Extra  \\
    \hline 
    C10 & ResNet20 & ReLU &88.24 &73.77 &62.05 &71.26 &76.48 &79.69 &76.46 \\
    C10 & ResNet20 & SiLU &87.53 &74.3 &63.32 &71.74 &77.33 &79.77 &76.58 \\
    C10 & ResNet20 & LA-SiLU &\textbf{88.52} &\textbf{75.31} &\textbf{63.81} &\textbf{73.11} &\textbf{78.23} &\textbf{80.92} &\textbf{77.6} \\
    \hline 
    C10 & ResNet32 & ReLU &\textbf{88.55} &74.48 &63.33 &71.92 &76.94 &80.33 &77.11 \\
    C10 & ResNet32 & SiLU  &87.27  &75.3  &64.78  &72.92  &78.55  &80.3  &77.32 \\
    C10 & ResNet32 & LA-SiLU &87.85 &\textbf{76.39} &\textbf{67.35} &\textbf{74.22} &\textbf{78.80} & \textbf{80.99} & \textbf{78.32} \\
    \hline 
    C10 & ResNet44 & ReLU &\textbf{88.58} &75.2 &64.32 &72.75 &77.87 &\textbf{80.67} &77.67 \\
    C10 & ResNet44 & SiLU &86.65 &75.11 &65.67 &72.73 &77.84 &79.85 &77.1 \\
    C10 & ResNet44 & LA-SiLU &86.88 &\textbf{76.73} &\textbf{69.53} &\textbf{74.98} &\textbf{78.51} &80.43 &\textbf{78.43} \\
    \hline 
    C100 & ResNet20 & ReLU &61.30 &44.04 &31.82 &44.39 &47.03 &48.86 &45.07 \\
    C100 & ResNet20 & SiLU &60.45 &\textbf{45.93} &\textbf{35.66} &\textbf{46.46} &48.46 &49.91 &\textbf{46.59} \\
    C100 & ResNet20 & LA-SiLU &\textbf{62.30} &45.30 &31.63 &45.90 &\textbf{48.72} &\textbf{50.65} &46.17 \\
    \hline 
    C100 & ResNet32 & ReLU &\textbf{63.21} &45.92 &33.23 &46.17 &49.26 &50.94 &46.84 \\
    C100 & ResNet32 & SiLU &60.04 &47.14 &\textbf{37.37} &\textbf{48.60} &49.57 &50.33 &47.41 \\
    C100 & ResNet32 & LA-SiLU &60.76 &\textbf{47.26} &36.63 &48.51 &\textbf{49.97} &\textbf{51.03} &\textbf{47.51} \\
    \hline 
    C100 & ResNet44 & ReLU &\textbf{64.18} &46.63 &33.92 &46.89 &\textbf{49.68} &\textbf{51.72} &\textbf{47.75} \\
    C100 & ResNet44 & SiLU &59.58 &\textbf{47.41} &\textbf{38.28} &48.69 &49.45 &50.61 &47.73 \\
    C100 & ResNet44 & LA-SiLU &60.17 &47.31 &38.10 &\textbf{48.82} &49.63 &50.26 &47.46 \\
\hline \hline
\end{tabular}
\end{table}

\section{H. LayerAct with no normalization or LN}
\label{app_layeract_with_LN}

In this section, we explore the cases when LA-SiLU is utilized with LN. For the further investigation, we first examine the relationship between LN and LayerAct, subsequently presenting experimental results on CIFAR datasets for two scenarios: (1) networks without any normalization methods, and (2) networks with LN. ReLU and SiLU were selected as baseline element-level activation functions, with ReLU being the most popular choice for CNNs, and SiLU being as the corresponding element-level activation function of LA-SiLU. 

For the networks incoporating LN, the normalization output of a LN layer is expressed as $n^{\hat{LN}}=\gamma \cdot \frac{y-\mu_{y}^{2}}{\sqrt{\sigma_{y}^{2}+\alpha}} + \beta$. This equation results in the output of SiLU $n^{\hat{LN}}s\left(n^{\hat{LN}}\right)$ and LA-SiLU $n^{\hat{LN}}s\left(n\right)$ becoming similar. When the learnable parameter $\gamma$ and $\beta$ are $1$ and $0$, respectively, the outputs $n^{\hat{LN}}s\left(n^{\hat{LN}}\right)$ and $n^{\hat{LN}}s\left(n\right)$ become identical, nullifying any distinct advantage offered by the LayerAct mechanism. Thus, the more the output of the LN layer approximates the normalized input of LayerAct's activation scale function (i.e., the more gain and bias approximate to $1$ and $0$, respectively), the more the benefits of LayerAct are reduced.

Nevertheless, the benefits of LayerAct functions, addressing the trade-off between two important activation properties and potentially exhibiting lower variance in robustness across samples, can still be partially retained when combined with LN that the learnable parameters $\gamma$ and $\beta$ are far from $1$ and $0$, respectively. This is because the layer-direction normalized input of LayerAct's activation scale function does not utilize an affine function. By not utilizing an affine function, we can ensure that $\left\Vert s\left( \hat{n} \right) \right\Vert$ of LayerAct functions remains small enough for robustness. When an affine function is used, the mean of the normalized vector $n$ can become large if the parameter $\beta$ is large. Such large $n$ leads to a large $\left\Vert s\left( \hat{n} \right) \right\Vert$, which does not ensure robustness. 

Additionally, LayerAct functions can allow a larger negative output compared to element-level activation functions, as they do not restrict large negative outputs. Furthermore, the robustness benefit of LayerAct becomes significant when the normalized outputs of LN (i.e. activation input) are mostly outside the saturation state. 

To investigate this, we conducted experiments on ResNets both without normalization and with LN. We followed the same experimental setting of the experiments in our main manuscript. The results reported in the tables are the average accuracies over 30 runs, each with different seeds for weight initialization. 

Table \ref{table_appendix_without_norm} presents the performance of ResNets without any normalization methods on both clean and noisy CIFAR datasets, with a learning rate set at 0.01 for stable training. The results demonstrate that networks with LA-SiLU outperformed those with ReLU and SiLU on noisy datasets. This indicates that the robustness of LayerAct functions is independent from the presence of normalization layers. Additionally, while networks with SiLU, the corresponding element-level activation function, experienced training instability and often exploded, those with LA-SiLU maintained stable training. This suggests that the LayerAct mechanism can contribute to more stable network training. However, in deeper networks such as ResNet32 and ResNet44, the performance with ReLU surpassed those with LA-SiLU. This outcome may be because of the more complex representation of LA-SiLU's output, pointing towards the necessity of a normalization layer to mitigate potential overfitting issues. 

Table \ref{table_appendix_with_LN} demonstrates the performance of ResNets with LN on both clean and noisy CIFAR10 and CIFAR100 datasets. It validated our concerns about the relationship between LN and LayerAct. On noisy CIFAR100, ResNet20 and ResNet44 with SiLU demonstrated similar or greater robustness than those with LA-SiLU. This suggests that the robustness benefits of LayerAct are diminished when utilized with LN. However, it is important to note that the benefits of LayerAct may still be partially preserved even when LN is applied. For clean and noisy CIFAR10, the results where similar with those in Table \ref{table_appendix_without_norm}, the networks with LA-SiLU demonstrated more robust inference compared to those with other activation functions on noisy datasets, while the performance of the deeper networks (ResNet32 and ResNet44) with ReLU where better on clean datasets than those with LA-SiLU. Furthermore, on clean datasets, the networks with LA-SiLU outperformed those with SiLU, which employs the same function (Logistic Sigmoid) for the activation scale with LA-SiLU. This indicates that the advantage of LA-SiLU, in addressing the trade-off, can enhance network performance.

Additionally, networks employing LayerAct functions with BN, which normalizes in a different dimension from that of LayerAct functions and remains the dominant normalization method for CNNs, perform effectively. In our experiments reported in the main manuscript, BN was employed for all networks, except for UNets, which do not utilize any normalization layer. The networks with LayerAct functions exhibited remarkable performance on both clean and noisy datasets. Consequently, despite the necessity for careful selection of the normalization method when using LayerAct functions, their robustness and improved performance when combined with BN make them an attractive choice for CNNs.

%%%%%%%%%%%%%%%%%%%%%%%%%%%%%%%%%%%%%%%%%%%%%%%%%%%%%%%%%%%%%%%%%%%%%%

\begin{table}[tb]
\caption{Classification performance on CIFAR and CIFAR-C datasets with SN. We report the average mean accuracy in the table. C10 and C100 denotes CIFAR10 and CIFAR100, respectively.}
\label{table_appendix_with_switch}
\centering
\begin{tabular}{@{}llllllllll@{}}
\hline \hline
    & & & \textbf{CIFAR} & \multicolumn{6}{c}{\bf CIFAR-C} \\
    Data & Model & Activation & Clean & Total & Noise & Blur & Digital & Weather & Extra  \\ 
    \hline 
    C10 & ResNet20 & ReLU &89.65 &72.40 &56.07 &71.35 &75.24 &80.43 &74.81 \\
    C10 & ResNet20 & SiLU &\textbf{90.60} &\textbf{74.17} &\textbf{57.70} &\textbf{72.94} &\textbf{77.57} &\textbf{82.21} &\textbf{76.33} \\
    C10 & ResNet20 & LA-SiLU &89.56 &72.43 &55.9 &71.05 &75.47 &80.87 &74.73 \\
    \hline 
    C10 & ResNet32 & ReLU &90.70 &73.90 &58.24 &72.84 &76.3 &81.79 &76.42 \\
    C10 & ResNet32 & SiLU &\textbf{90.79} &\textbf{74.65} &\textbf{58.78} &\textbf{73.39} &\textbf{77.64} &\textbf{82.57} &\textbf{76.91} \\
    C10 & ResNet32 & LA-SiLU &89.94 &72.72 &56.15 &71.07 &75.52 &81.48 &75.23 \\
    \hline 
    C10 & ResNet44 & ReLU &\textbf{91.40} &\textbf{74.65} &\textbf{58.10} &\textbf{74.01} &\textbf{77.18} &\textbf{82.81} &\textbf{76.99} \\
    C10 & ResNet44 & SiLU &74.48 &61.70 &49.50 &60.58 &64.11 &67.87 &63.41 \\
    C10 & ResNet44 & LA-SiLU &89.36 &72.02 &54.85 &70.59 &75.07 &81.05 &74.24 \\
    \hline 
    C100 & ResNet20 & ReLU &57.36 &37.01 &20.61 &38.43 &39.75 &43.56 &38.59 \\
    C100 & ResNet20 & SiLU &\textbf{64.55} &\textbf{42.96} &\textbf{25.13} &\textbf{43.91} &\textbf{46.35} &\textbf{50.42} &\textbf{44.55} \\
    C100 & ResNet20 & LA-SiLU &63.88 &41.76 &23.23 &42.83 &45.28 &49.43 &43.4 \\
    \hline 
    C100 & ResNet32 & ReLU &60.19 &38.81 &21.89 &40.02 &41.30 &46.00 &40.61 \\
    C100 & ResNet32 & SiLU &\textbf{64.35} &\textbf{42.45} &\textbf{25.00} &43.08 &\textbf{45.86} &49.92 &\textbf{44.00} \\
    C100 & ResNet32 & LA-SiLU &64.05 &42.27 &23.72 &\textbf{43.46} &45.50 &\textbf{50.16} &43.89 \\
    \hline 
    C100 & ResNet44 & ReLU &61.64 &39.93 &22.64 &41.03 &42.47 &47.54 &41.67 \\
    C100 & ResNet44 & SiLU &44.8 &29.57 &18.11 &29.96 &31.98 &34.21 &30.73 \\
    C100 & ResNet44 & LA-SiLU &\textbf{62.99} &\textbf{41.91} &\textbf{22.86} &\textbf{43.06} &\textbf{46.02} &\textbf{49.79} &\textbf{43.06} \\
\hline \hline
\end{tabular}
\end{table}

%%%%%%%%%%%%%%%%%%%%%%%%%%%%%%%%%%%%%%%%%%%%%%%%%%%%%%%%%%%%%%%%%%%%%%

\section{I. LayerAct with Other Normalization Methods.}

To investigate the relationship between LayerAct and various normalization methods, we conducted experiments on ResNets with Switchable Normalization (SN) \cite{luo2019switchable}, Instance enhancement batch normalization (IEBN) \cite{liang2020instance}, and Decorrelated Batch Normalization (DBN) \cite{Huang2018decorr}. We used the same experiment setting with those of our main manuscript. We report the average accuracy over 10 runs for the experiments.

Analysis of the experimental results revealed that LayerAct functions are not effectively compatible with normalizations that can cause a large variance in the channel means, such as SN and IEBN. This incompatibility arises because LayerAct is more sensitive to such large variance between channel means, which causes channels with smaller means to be more likely to become inactivated compared to those with larger means.

Table \ref{table_appendix_with_switch} demonstrates that utilizing LayerAct with SN is not effective. This ineffectiveness arises because SN is a combination of BN, LN, and Instance normalization (IN) \cite{ulyanov2016instance}, while the benefit of LayerAct functions diminish when preceded by LN. Additionally, LayerAct functions are more sensitive to the presence of similar channel characteristics across samples compared to element-level activation functions because the channel dimension is incorporated into normalizing dimension of LayerAct's activation scale input. IN aggressively normalizes the mean and variance across channels towards uniformity. Therefore, given the composite normalization strategy of SN, we expect that the effect of IN will lead to a homogenization of channel characteristics, thus undermining the efficiency of LayerAct functions. 

\begin{table}[ht]
\caption{Classification performance on CIFAR and CIFAR-C datasets with IEBN. We report the average mean accuracy in the table. C10 and C100 denotes CIFAR10 and CIFAR100, respectively.}
\label{table_appendix_with_IEBN}
\centering
\begin{tabular}{@{}llllllllll@{}}
\hline \hline
    & & & \textbf{CIFAR} & \multicolumn{6}{c}{\bf CIFAR-C} \\
    Data & Model & Activation & Clean & Total & Noise & Blur & Digital & Weather & Extra  \\ 
    \hline 
    C10 & ResNet20 & ReLU &91.50 &70.12 &52.49 &66.73 &73.44 &79.64 &73.89 \\
    C10 & ResNet20 & SiLU &91.44 &68.98 &49.39 &66.26 &72.29 &79.00 &73.06 \\
    C10 & ResNet20 & LA-SiLU &\textbf{91.63} & \textbf{70.64} & \textbf{52.70} & \textbf{67.30} & \textbf{74.20} & \textbf{79.87} & \textbf{74.65} \\
    \hline 
    C10 & ResNet32 & ReLU & \textbf{92.54} &71.67 &54.06 &\textbf{68.82} &74.82 &81.12 &75.11 \\
    C10 & ResNet32 & SiLU &91.85 &71.07 &53.39 &68.26 &73.90 &80.50 &74.90 \\
    C10 & ResNet32 & LA-SiLU &92.36 &\textbf{71.77} &\textbf{54.19} &68.19 &\textbf{75.32} &\textbf{81.15} &\textbf{75.60} \\
    \hline 
    C10 & ResNet44 & ReLU &\textbf{92.78} &72.18 &55.20 &68.83 &75.61 &81.59 &75.43 \\
    C10 & ResNet44 & SiLU &92.08 &71.23 &52.79 &68.54 &74.27 &80.95 &74.99 \\
    C10 & ResNet44 & LA-SiLU &92.50 &\textbf{73.01} &\textbf{56.97} &\textbf{69.18} &\textbf{76.03} &\textbf{82.34} &\textbf{76.5} \\
    \hline 
    C100 & ResNet20 & ReLU &66.62 &\textbf{41.97} &\textbf{22.81} &\textbf{42.51} &45.07 &50.30 &44.37 \\
    C100 & ResNet20 & SiLU &66.19 &40.88 &21.14 &41.08 &44.29 &49.68 &43.28 \\
    C100 & ResNet20 & LA-SiLU &\textbf{66.73} &41.59 &21.31 &41.98 &\textbf{45.26} &\textbf{50.53} &\textbf{43.82} \\
    \hline 
    C100 & ResNet32 & ReLU &\textbf{68.17} &43.49 &23.81 &\textbf{43.89} &46.76 &52.30 &\textbf{45.78} \\
    C100 & ResNet32 & SiLU &67.43 &42.30 &23.15 &42.38 &45.33 &51.14 &44.68 \\
    C100 & ResNet32 & LA-SiLU &67.97 &\textbf{43.56} &\textbf{24.49} &43.39 &\textbf{46.94} &\textbf{52.47} &45.76 \\
    \hline 
    C100 & ResNet44 & ReLU &\textbf{69.47} &44.73 &25.74 &44.77 &47.77 &53.51 &47.12 \\
    C100 & ResNet44 & SiLU &68.18 &43.47 &24.76 &43.37 &46.46 &52.22 &45.89 \\
    C100 & ResNet44 & LA-SiLU &68.41 &\textbf{45.29} &\textbf{26.89} &\textbf{45.15} &\textbf{48.67} &\textbf{54.03} &\textbf{47.13} \\
\hline \hline
\end{tabular}
\end{table}

Table \ref{table_appendix_with_IEBN} demonstrates the performance of networks with IEBN. With the exception of ResNet20 on CIFAR100, networks with LA-SiLU demonstrated enhanced performance on noisy datasets when compared to their counterparts utilizing ReLU and SiLU. Conversely, ReLU outperformed LA-SiLU in ResNet32 and ResNet44 models. Nonetheless, it is important to highlight that LA-SiLU consistently surpassed SiLU, which utilizes the same activation scale function, across all tested scenarios. These results imply that the mechanism of LayerAct holds promise for enhancing efficiency. It is also important to consider careful consideration and integration of network complexity, particularly due to the interplay between normalization and activation functions, are essential, given that the scale function of ReLU is considerably simpler compared to sigmoid, the scale function of LA-SiLU and SiLU.

\begin{table}[ht]
\caption{Classification performance on CIFAR and CIFAR-C with DBN. We report the average mean accuracy in the table. C10 and C100 denotes CIFAR10 and CIFAR100, respectively.}
\label{table_appendix_with_DBN}
\centering
\begin{tabular}{@{}llllllllll@{}}
\hline \hline
    & & & \textbf{CIFAR} & \multicolumn{6}{c}{\bf CIFAR-C} \\
    Data & Model & Activation & Clean & Total & Noise & Blur & Digital & Weather & Extra  \\ 
    \hline 
    C10 & ResNet20 & ReLU &90.8 &65.52 &35.87 &65.43 &71.27 &78.68 &68.93 \\
    C10 & ResNet20 & SiLU &87.69 &63.55 &38.16 &61.21 &70.24 &74.77 &67.01  \\
    C10 & ResNet20 & LA-SiLU &\textbf{91.72} &\textbf{69.36} &\textbf{45.68} &\textbf{67.81} &\textbf{73.41} &\textbf{80.6} &\textbf{73.37}  \\
    \hline 
    C10 & ResNet32 & ReLU &91.85 &68.1 &37.15 &68.38 &\textbf{75.07} &81.04 &71.1  \\
    C10 & ResNet32 & SiLU &92.22 &68.3 &40.01 &67.24 &73.98 &81.44 &71.77  \\
    C10 & ResNet32 & LA-SiLU &\textbf{92.27} &\textbf{70.36} &\textbf{45.62} &\textbf{69.32} &74.58 &\textbf{81.96} &\textbf{74.14}  \\ 
    \hline 
    C10 & ResNet44 & ReLU &92.35 &69.79 &39.25 &\textbf{70.48} &76.61 &\textbf{82.31} &72.67  \\
    C10 & ResNet44 & SiLU &\textbf{92.39} &69.48 &42.19 &68.97 &74.63 &81.63 &73.13  \\
    C10 & ResNet44 & LA-SiLU &92.36 &\textbf{70.17} &\textbf{44.0} &69.35 &\textbf{74.63} &82.22 &\textbf{74.11}  \\
    \hline 
    C100 & ResNet20 & ReLU &\textbf{58.38} &\textbf{34.19} &\textbf{13.48} &\textbf{34.26} &\textbf{38.28} &\textbf{43.67} &\textbf{36.07}  \\
    C100 & ResNet20 & SiLU &57.82 &32.13 &13.41 &31.57 &35.78 &40.99 &34.22  \\
    C100 & ResNet20 & LA-SiLU &58.19 &32.24 &14.32 &30.95 &35.11 &41.4 &34.96  \\
    \hline 
    C100 & ResNet32 & ReLU &54.06 &30.29 &14.16 &28.67 &34.09 &38.34 &32.14  \\
    C100 & ResNet32 & SiLU &62.0 &35.26 &14.99 &34.41 &39.35 &45.01 &37.46  \\
    C100 & ResNet32 & LA-SiLU &\textbf{65.02} &\textbf{38.75} &\textbf{19.18} &\textbf{37.84} &\textbf{41.81} &\textbf{48.66} &\textbf{41.36}  \\
    \hline 
    C100 & ResNet44 & ReLU &60.32 &34.64 &14.96 &33.68 &39.03 &44.1 &36.51  \\
    C100 & ResNet44 & SiLU &64.67 &37.58 &15.81 &37.63 &42.01 &47.2 &39.83  \\
    C100 & ResNet44 & LA-SiLU &\textbf{67.59} &\textbf{42.44} &\textbf{21.03} &\textbf{42.54} &\textbf{46.52} &\textbf{52.15} &\textbf{44.6}  \\
\hline \hline
\end{tabular}
\end{table}

Table \ref{table_appendix_with_DBN} demonstrates the performance of networks with DBN. Except for ResNet20 on CIFAR100, networks with LA-SiLU showed similar or improved performance on clean datasets, and outstanding performance on noisy datasets compared to those employing ReLU and SiLU. The results of these experiments highlight the potential applicability of LayerAct functions in conjunction with advanced batch-direction normalization methods.

%%%%%%%%%%%%%%%%%%%%%%%%%%%%%%%%%%%%%%%%%%%%%%%%%%%%%%%%%%%%%%%%%%%%%%

\begin{table}[ht]]
\caption{Classification performance on the clean CIFAR10 and CIFAR100}
\label{table_appendix_clean_cifar}
\centering
\begin{tabular}{@{}lllllll@{}}
\hline \hline
    & \multicolumn{3}{c}{\bf CIFAR10} & \multicolumn{3}{c}{\bf CIFAR100} \\
    Activation & ResNet20 & ResNet32 & ResNet44 & ResNet20 & ResNet32 & ResNet44  \\ 
    \hline 
    ReLU & 91.29 & 92.03 & 92.03 & 65.92 & 67.04 & 68.02 \\
    LReLU & 91.31 & 92.03 & 92.03 & 65.88 & \textbf{67.37} & 67.96 \\
    PReLU & 90.82 & 92.03 & - & 64.00 & 66.35 & 67.68 \\
    SiLU & 91.45 & 92.17 & 92.18 & 65.89 & 67.22 & 67.71 \\
    HardSiLU & 91.09 & 91.77 & 91.42 & 65.19 & 66.49 & 66.38 \\
    Mish & 91.48 & \textbf{92.21} & 92.30 & 65.85 & 67.18 & 68.06 \\
    GELU & \textbf{91.50} & \textbf{\underline{92.25}} & \textbf{92.22} & 65.84 & 67.30 & \textbf{\underline{68.19}} \\
    ELU & 91.04 & 91.61 & 91.68 & \textbf{66.24} & 67.01 & 67.55 \\
    \hline 
    LA-SiLU & \textbf{\underline{91.60}} & 92.20 & \textbf{\underline{92.36}} & \textbf{\underline{66.39}} & \textbf{\underline{67.74}} & \textbf{68.07} \\
    LA-HardSiLU & 91.21 & 91.68 & 91.36 & 66.16 & 66.63 & 65.51 \\
\hline \hline
\end{tabular}
\end{table}

%%%%%%%%%%%%%%%%%%%%%%%%%%%%%%%%%%%%%%%%%%%%%%%%%%%%%%%%%%%%%%%%%%%%%%

\section{J. Additional Tables}

Table \ref{table_appendix_clean_cifar} shows the classification performance of networks on clean CIFAR10 and CIFAR100 datasets. LA-SiLU outperformed element-level activation functions in four out of six cases. Specifically, LA-SiLU always demonstrated superior performance compared to SiLU. 

\begin{table}[ht]]
\caption{Classification performance of ResNet32 and ResNet44 on noisy CIFAR10 and CIFAR100. We report the average mean accuracy over 30 runs.}
\label{table_appendix_c_32_44}
\centering
\begin{tabular}{@{}lllllllll@{}}
\hline \hline
    Data & Model & Activation & Total & Noise & Blur & Digital & Weather & Extra  \\ 
    \hline 
    CIFAR10-C & ResNet32 & ReLU & 72.00 & 53.07 & 67.62 & 74.75 & 80.44 & 74.41 \\
    CIFAR10-C & ResNet32 & LReLU & 72.01 & 52.66 & \textbf{67.86} & \textbf{74.77} & 80.42 & 74.5 \\
    CIFAR10-C & ResNet32 & PReLU & 71.7 & 52.82 & 67.06 & 74.72 & 79.90 & 74.17 \\
    CIFAR10-C & ResNet32 & SiLU & 71.7 & 52.37 & 67.21 & 74.08 & 80.51 & 74.38 \\
    CIFAR10-C & ResNet32 & HardSiLU & 71.32 & 52.75 & 66.75 & 73.39 & 79.66 & 74.28 \\
    CIFAR10-C & ResNet32 & Mish & 71.96 & 53.09 & 67.42 & 74.3 & 80.57 & 74.64 \\
    CIFAR10-C & ResNet32 & GELU & 71.64 & 52.44 & 67.18 & 74.11 & 80.26 & 74.26 \\
    CIFAR10-C & ResNet32 & LA-SiLU & \textbf{\underline{72.8}} & \textbf{54.02} & \textbf{\underline{68.42}} & \textbf{\underline{75.13}} & \textbf{\underline{81.36}} & \textbf{75.51} \\
    CIFAR10-C & ResNet32 & LA-HardSiLU & \textbf{72.6} & \textbf{\underline{55.36}} & 67.71 & 74.70 & \textbf{80.53} & \textbf{\underline{75.62}} \\
    \hline 
    CIFAR10-C & ResNet44 & ReLU & \textbf{\underline{73.71}} & \textbf{56.39} & \textbf{70.03} & \textbf{76.05} & 81.27 & 75.92 \\
    CIFAR10-C & ResNet44 & LReLU & \textbf{73.69} & 56.03 & \textbf{\underline{70.10}} & \textbf{\underline{76.09}} & \textbf{81.43} & 75.81 \\
    CIFAR10-C & ResNet44 & SiLU & 72.45 & 53.12 & 68.64 & 74.80 & 80.88 & 75.06 \\
    CIFAR10-C & ResNet44 & HardSiLU & 72.63 & 55.51 & 68.72 & 74.73 & 79.86 & 75.34 \\
    CIFAR10-C & ResNet44 & Mish & 72.79 & 53.74 & 68.86 & 75.22 & 81.18 & 75.3 \\
    CIFAR10-C & ResNet44 & GELU & 72.82 & 54.65 & 68.69 & 75.26 & 80.85 & 75.24 \\
    CIFAR10-C & ResNet44 & LA-SiLU & 73.5 & 55.29 & 69.14 & 75.73 & \textbf{\underline{81.91}} & \textbf{76.19} \\
    CIFAR10-C & ResNet44 & LA-HardSiLU & 73.33 & \textbf{\underline{57.45}} & 68.48 & 75.30 & 80.64 & \textbf{\underline{76.30}} \\
    \hline 
    CIFAR100-C & ResNet32 & ReLU & 43.51 & 24.10 & 41.99 & 45.81 & 50.58 & 44.32 \\
    CIFAR100-C & ResNet32 & LReLU & 43.58 & 23.8 & 42.12 & 45.94 & 50.73 & 44.42 \\
    CIFAR100-C & ResNet32 & PReLU & 42.44 & 23.72 & 40.31 & 44.81 & 49.42 & 43.27 \\
    CIFAR100-C & ResNet32 & SiLU & 42.94 & 23.06 & 41.27 & 45.01 & 50.31 & 44.02 \\
    CIFAR100-C & ResNet32 & HardSiLU & 42.67 & 23.64 & 40.87 & 44.87 & 49.54 & 43.71 \\
    CIFAR100-C & ResNet32 & Mish & 42.95 & 22.69 & 41.53 & 45.05 & 50.37 & 43.97 \\
    CIFAR100-C & ResNet32 & GELU & 43.00 & 23.15 & 41.45 & 45.21 & 50.21 & 43.94 \\
    CIFAR100-C & ResNet32 & LA-SiLU & \textbf{44.6} & \textbf{24.16} & \textbf{43.58} & \textbf{46.74} & \textbf{\underline{52.11}} & \textbf{45.51} \\
    CIFAR100-C & ResNet32 & LA-HardSiLU & \textbf{\underline{44.86}} & \textbf{\underline{25.98}} & \textbf{\underline{43.83}} & \textbf{\underline{47.02}} & \textbf{51.50} & \textbf{\underline{45.81}} \\
    \hline 
    CIFAR100-C & ResNet44 & ReLU & 44.77 & 25.45 & 43.05 & 47.33 & 51.94 & 45.44 \\
    CIFAR100-C & ResNet44 & LReLU & 44.8 & 25.57 & 43.26 & 47.17 & 51.91 & 45.5 \\
    CIFAR100-C & ResNet44 & PReLU & 44.31 & 25.78 & 42.19 & 46.7 & 51.44 & 44.97 \\
    CIFAR100-C & ResNet44 & SiLU & 44.04 & 24.52 & 42.61 & 46.16 & 51.08 & 45.01 \\
    CIFAR100-C & ResNet44 & HardSiLU & 44.11 & 26.22 & 42.77 & 46.29 & 50.32 & 44.9 \\
    CIFAR100-C & ResNet44 & Mish & 44.14 & 24.18 & 42.69 & 46.36 & 51.37 & 45.11 \\
    CIFAR100-C & ResNet44 & GELU & 43.93 & 24.39 & 42.29 & 46.15 & 51.02 & 44.85 \\
    CIFAR100-C & ResNet44 & LA-SiLU & \textbf{46.12} & \textbf{26.68} & \textbf{44.94} & \textbf{48.32} & \textbf{\underline{53.44}} & \textbf{46.89} \\
    CIFAR100-C & ResNet44 & LA-HardSiLU & \textbf{\underline{46.83}} & \textbf{\underline{30.85}} & \textbf{\underline{45.83}} & \textbf{\underline{48.93}} & \textbf{52.5} & \textbf{\underline{47.35}} \\
\hline \hline
\end{tabular}
\end{table}

\begin{table}[ht]
\caption{Standard deviation of ResNet20s' accuracy on CIFAR10 and CIFAR10-C with different activation functions. We average in the table.}
\label{table_appendix_std_CIFAR10_resnet20}
\centering
\begin{tabular}{@{}llllllll@{}}
\hline \hline
Activation & Clean & Total & Noise & Blur & Digital & Weather & Extra \\
    \hline 
ReLU & 0.2120 & 1.3171 & 2.4428 & 1.5145 & 0.9736 & 0.7231 & 1.2127 \\
LReLU & 0.2635 & 1.2630 & 2.2750 & 1.4947 & 0.8948 & 0.6938 & 1.2099 \\
PReLU & 0.2323 & 1.4176 & 2.4859 & 1.4106 & 1.3038 & 0.8178 & 1.3369 \\
SiLU & 0.2057 & 1.1067 & 1.7695 & 1.3471 & 0.8738 & 0.7012 & 1.0076 \\
HardSiLU & 0.2287 & 1.2052 & 2.3675 & 1.4039 & 0.8074 & 0.6465 & 1.0914 \\
MISH & 0.2302 & 1.1507 & 2.1373 & 1.3116 & 0.7921 & 0.6335 & 1.1255 \\
GELU & 0.1868 & 1.1358 & 1.8216 & 1.5201 & 0.8537 & 0.5251 & 1.1298 \\
ELU & 0.1742 & 1.1673 & 2.2787 & 1.3572 & 0.7776 & 0.5961 & 1.1047 \\
    \hline 
LA-SiLU & 0.1645 & 1.1800 & 1.8781 & 1.4681 & 0.9738 & 0.7475 & 1.0069 \\
LA-HardSiLU & 0.2051 & 1.1765 & 2.3889 & 1.1128 & 0.9142 & 0.7081 & 1.0616 \\
\hline \hline
\end{tabular}
\end{table}

\begin{table}[ht]
\caption{Standard deviation of ResNet32s' accuracy on CIFAR10 and CIFAR10-C with different activation functions. We average in the table.}
\label{table_appendix_std_CIFAR10_resnet32}
\centering
\begin{tabular}{@{}llllllll@{}}
\hline \hline
Activation & Clean & Total & Noise & Blur & Digital & Weather & Extra \\
    \hline 
ReLU & 0.3880 & 1.5955 & 2.8978 & 1.7971 & 1.3270 & 0.9254 & 1.3556 \\
LReLU & 0.3356 & 1.7599 & 3.0043 & 2.1394 & 1.4202 & 1.0200 & 1.5270 \\
PReLU & 0.2809 & 1.6875 & 2.6927 & 1.8571 & 1.6496 & 0.9261 & 1.5634 \\
SiLU & 0.2188 & 1.3848 & 2.7176 & 1.7485 & 0.8207 & 0.7220 & 1.2483 \\
HardSiLU & 0.2200 & 1.2341 & 2.0973 & 1.5508 & 0.8625 & 0.8104 & 1.0653 \\
MISH & 0.2497 & 1.2869 & 2.1864 & 1.7262 & 0.8797 & 0.6475 & 1.2194 \\
GELU & 0.1568 & 1.2598 & 2.3925 & 1.4798 & 0.8745 & 0.6164 & 1.2188 \\
ELU & 0.1800 & 1.2041 & 1.8703 & 1.5525 & 0.9373 & 0.7095 & 1.1177 \\
    \hline 
LA-SiLU & 0.1898 & 1.2232 & 1.9876 & 1.4006 & 1.1328 & 0.6804 & 1.1056 \\
LA-HardSiLU & 0.3023 & 1.4008 & 2.2405 & 1.7160 & 1.1416 & 0.8028 & 1.3129 \\
\hline \hline
\end{tabular}
\end{table}

\begin{table}[ht]
\caption{Standard deviation of ResNet44s' accuracy on CIFAR10 and CIFAR10-C with different activation functions. We average in the table.}
\label{table_appendix_std_CIFAR10_resnet44}
\centering
\begin{tabular}{@{}llllllll@{}}
\hline \hline
Activation & Clean & Total & Noise & Blur & Digital & Weather & Extra \\
    \hline 
ReLU & 0.6249 & 2.1269 & 3.5833 & 2.4709 & 1.8301 & 1.2399 & 1.8742 \\
LReLU & 0.4437 & 1.9929 & 3.3751 & 2.3684 & 1.6750 & 1.1650 & 1.7264 \\
PReLU & 14.5207 & 11.3037 & 8.8671 & 10.8069 & 11.8174 & 12.5987 & 11.8192 \\
SiLU & 0.2435 & 1.4765 & 2.6747 & 1.7969 & 1.0507 & 0.8144 & 1.3453 \\
HardSiLU & 0.4192 & 1.7214 & 3.2993 & 1.8791 & 1.3047 & 1.0844 & 1.4341 \\
MISH & 0.4472 & 2.1393 & 3.7470 & 2.5737 & 1.6741 & 1.2354 & 1.8683 \\
GELU & 0.4696 & 1.6769 & 3.1032 & 1.8889 & 1.2754 & 0.8846 & 1.5892 \\
ELU & 0.2073 & 1.4907 & 2.7899 & 1.7389 & 1.1132 & 0.7598 & 1.3765 \\
    \hline 
LA-SiLU & 0.2719 & 1.3634 & 2.5109 & 1.5678 & 1.0625 & 0.7445 & 1.2180 \\
LA-HardSiLU & 0.2033 & 1.5738 & 2.8731 & 1.7762 & 1.3225 & 0.8906 & 1.3316 \\
\hline \hline
\end{tabular}
\end{table}

\begin{table}[ht]
\caption{Standard deviation of ResNet20s' accuracy on CIFAR100 and CIFAR100-C with different activation functions. We average in the table.}
\label{table_appendix_std_CIFAR100_resnet20}
\centering
\begin{tabular}{@{}llllllll@{}}
\hline \hline
Activation & Clean & Total & Noise & Blur & Digital & Weather & Extra \\
    \hline 
ReLU & 0.2937 & 0.8120 & 1.0754 & 0.9131 & 0.7956 & 0.5861 & 0.7558 \\
LReLU & 0.3233 & 0.8633 & 1.0476 & 0.8994 & 0.7943 & 0.7862 & 0.8353 \\
PReLU & 0.4435 & 0.9101 & 1.2044 & 1.0421 & 0.8121 & 0.7970 & 0.7682 \\
SiLU & 0.4147 & 0.7665 & 1.0922 & 0.8286 & 0.6642 & 0.5197 & 0.8095 \\
HardSiLU & 0.3903 & 0.8104 & 1.1921 & 0.8172 & 0.6649 & 0.6375 & 0.8356 \\
MISH & 0.3269 & 0.7400 & 1.0435 & 0.6064 & 0.7377 & 0.6462 & 0.7420 \\
GELU & 0.3866 & 0.8059 & 0.9036 & 0.9290 & 0.8281 & 0.6220 & 0.7714 \\
ELU & 0.3618 & 0.7367 & 1.0898 & 0.7222 & 0.6868 & 0.5689 & 0.7044 \\
    \hline 
LA-SiLU & 0.3493 & 0.8537 & 1.3183 & 0.8409 & 0.7137 & 0.6806 & 0.8313 \\
LA-HardSiLU & 0.4056 & 0.8751 & 1.2555 & 0.8963 & 0.7188 & 0.7466 & 0.8536 \\
\hline \hline
\end{tabular}
\end{table}

\begin{table}[ht]
\caption{Standard deviation of ResNet32s' accuracy on CIFAR100 and CIFAR100-C with different activation functions. We average in the table.}
\label{table_appendix_std_CIFAR100_resnet32}
\centering
\begin{tabular}{@{}llllllll@{}}
\hline \hline
Activation & Clean & Total & Noise & Blur & Digital & Weather & Extra \\
    \hline 
ReLU & 0.5187 & 1.0671 & 1.3930 & 1.1459 & 1.0240 & 0.8663 & 0.9876 \\
LReLU & 0.3364 & 0.9092 & 1.3616 & 0.8028 & 0.9881 & 0.6881 & 0.8184 \\
PReLU & 0.5045 & 0.9674 & 1.2581 & 1.0727 & 0.8316 & 0.8351 & 0.9123 \\
SiLU & 0.5053 & 1.0087 & 1.5275 & 1.0608 & 0.8957 & 0.6965 & 0.9926 \\
HardSiLU & 0.5325 & 0.9114 & 0.9530 & 1.0173 & 0.9229 & 0.9203 & 0.7538 \\
MISH & 0.4820 & 0.9148 & 1.2750 & 0.9058 & 0.8194 & 0.7269 & 0.9369 \\
GELU & 0.4418 & 0.8121 & 1.0506 & 0.9158 & 0.7070 & 0.6675 & 0.7790 \\
ELU & 1.4877 & 1.3180 & 1.1907 & 1.3131 & 1.3876 & 1.4028 & 1.2640 \\
    \hline 
LA-SiLU & 0.3776 & 0.8765 & 1.3721 & 0.8750 & 0.6403 & 0.6864 & 0.9326 \\
LA-HardSiLU & 0.4486 & 0.9763 & 1.5039 & 1.0204 & 0.8062 & 0.7838 & 0.8988 \\
\hline \hline
\end{tabular}
\end{table}

\begin{table}[ht]
\caption{Standard deviation of ResNet44s' accuracy on CIFAR100 and CIFAR100-C with different activation functions. We average in the table.}
\label{table_appendix_std_CIFAR100_resnet44}
\centering
\begin{tabular}{@{}llllllll@{}}
\hline \hline
Activation & Clean & Total & Noise & Blur & Digital & Weather & Extra \\
    \hline 
ReLU & 0.5652 & 1.1011 & 1.5926 & 1.1677 & 0.9634 & 0.9098 & 0.9947 \\
LReLU & 0.6872 & 1.2173 & 1.8513 & 1.1183 & 1.1814 & 0.9895 & 1.1046 \\
PReLU & 0.7084 & 1.3226 & 1.8320 & 1.4481 & 1.0939 & 1.2049 & 1.1616 \\
SiLU & 0.5204 & 0.9563 & 1.2746 & 0.9791 & 0.8475 & 0.8444 & 0.9157 \\
HardSiLU & 0.9417 & 1.4808 & 1.6740 & 1.5590 & 1.5011 & 1.3818 & 1.3362 \\
MISH & 0.5119 & 0.9386 & 1.3795 & 0.9206 & 0.7863 & 0.7879 & 0.9291 \\
GELU & 0.6182 & 0.9809 & 1.3545 & 0.9600 & 0.9044 & 0.8583 & 0.9207 \\
ELU & 0.4118 & 1.1195 & 1.6396 & 1.1471 & 0.9781 & 0.8948 & 1.0679 \\
    \hline 
LA-SiLU & 0.4366 & 1.1648 & 1.5511 & 1.2528 & 1.1577 & 0.9353 & 1.0238 \\
LA-HardSiLU & 0.7955 & 1.8629 & 2.9063 & 1.8458 & 1.6924 & 1.5680 & 1.5630 \\
\hline \hline
\end{tabular}
\end{table}

Table \ref{table_appendix_c_32_44} illustrates the classification performance of ResNet32 and ResNet44 on CIFAR10-C and CIFAR100-C datasets. We do not report the results for ResNet44 with PReLU on CIFAR10 due to network instability during training. LayerAct functions outperformed the element-level activation functions in all but one instance, specifically ResNet44 on CIFAR10-C. However, it is noteworthy that LA-SiLU and LA-HardSiLU demonstrated much robust inference compared to their corresponding element-level activation functions, SiLU and HardSiLU, in all cases. Tables \ref{table_appendix_std_CIFAR10_resnet20}, \ref{table_appendix_std_CIFAR10_resnet32}, \ref{table_appendix_std_CIFAR10_resnet44}, \ref{table_appendix_std_CIFAR100_resnet20}, \ref{table_appendix_std_CIFAR100_resnet32}, and \ref{table_appendix_std_CIFAR100_resnet44} demonstrate the standard deviation of the networks on the CIFAR and CIFAR-C dataset. 

\begin{table}[ht!]
\caption{Statistical significance test of LA-SiLU on CIFAR10 dataset.}
\label{table_appendix_stat_cifar10}
\centering
\begin{tabular}{@{}lllllllll@{}}
\hline \hline
    \multicolumn{9}{c}{\bf LA-SiLU} \\
     & ReLU & LReLU & PReLU & SiLU & HardSiLU & Mish & GELU & ELU \\
    \hline  
    RN20 & $<1e^{-3}$ & $<1e^{-3}$ & $<1e^{-3}$ & 0.002 & $<1e^{-3}$ & 0.011 & 0.016 & $<1e^{-3}$ \\
    RN32 & 0.02 & 0.011 & 0.005 & 0.331 & $<1e^{-3}$ & 0.422 & 0.125 & $<1e^{-3}$ \\
    RN44 & 0.014 & 0.001 & - & 0.007 & $<1e^{-3}$ & 0.479 & 0.094 & $<1e^{-3}$ \\
\hline \hline
\end{tabular}
\end{table}

\begin{table}[ht!]
\caption{Statistical significance test of LA-SiLU on CIFAR100 dataset.}
\label{table_appendix_stat_cifar100}
\centering
\begin{tabular}{@{}lllllllll@{}}
\hline \hline
    \multicolumn{9}{c}{\bf LA-SiLU} \\
     & ReLU & LReLU & PReLU & SiLU & HardSiLU & Mish & GELU & ELU \\
    \hline  
    RN20 & $<1e^{-3}$ & $<1e^{-3}$ & $<1e^{-3}$ & $<1e^{-3}$ & $<1e^{-3}$ & $<1e^{-3}$ & $<1e^{-3}$ & 0.065 \\
    RN32 & $<1e^{-3}$ & $<1e^{-3}$ & $<1e^{-3}$ & $<1e^{-3}$ & $<1e^{-3}$ & $<1e^{-3}$ & $<1e^{-3}$ & $<1e^{-3}$ \\
    RN44 & 0.357 & 0.244 & 0.008 & 0.006 & $<1e^{-3}$ & 0.476 & 0.207 & $<1e^{-3}$ \\
\hline \hline
\end{tabular}
\end{table}

\begin{table}[ht!]
\caption{Statistical significance test of LayerAct functions on CIFAR10-C dataset.}
\label{table_appendix_stat_cifar10c}
\centering
\begin{tabular}{@{}lllllllll@{}}
\hline \hline
    \multicolumn{9}{c}{\bf LA-SiLU} \\
     & ReLU & LReLU & PReLU & SiLU & HardSiLU & Mish & GELU & ELU \\
    \hline  
    RN20 & $<1e^{-3}$ & $<1e^{-3}$ & $<1e^{-3}$ & $<1e^{-3}$ & $<1e^{-3}$ & $<1e^{-3}$ & $<1e^{-3}$ & $<1e^{-3}$ \\
    RN32 & $<1e^{-3}$ & $<1e^{-3}$ & $<1e^{-3}$ & $<1e^{-3}$ & $<1e^{-3}$ & $<1e^{-3}$ & $<1e^{-3}$ & $<1e^{-3}$ \\
    RN44 & 0.399 & 0.233 & - & $<1e^{-3}$ & 0.001 & 0.002 & 0.003 & $<1e^{-3}$ \\
    \hline  
    \\
    \hline  
    \multicolumn{9}{c}{\bf LA-HardSiLU} \\ 
    & ReLU & LReLU & PReLU & SiLU & HardSiLU & Mish & GELU & ELU \\
    \hline  
    RN20 & $<1e^{-3}$ & $<1e^{-3}$ & $<1e^{-3}$ & $<1e^{-3}$ & $<1e^{-3}$ & $<1e^{-3}$ & $<1e^{-3}$ & $<1e^{-3}$ \\
    RN32 & 0.007 & 0.014 & $<1e^{-3}$ & $<1e^{-3}$ & $<1e^{-3}$ & $<1e^{-3}$ & $<1e^{-3}$ & $<1e^{-3}$ \\
    RN44 & 0.18 & 0.138 & - & $<1e^{-3}$ & 0.008 & 0.01 & 0.018 & $<1e^{-3}$ \\
\hline \hline
\end{tabular}
\end{table}

\begin{table}[ht!]
\caption{Statistical significance test of LayerAct functions on CIFAR100-C dataset.}
\label{table_appendix_stat_cifar100c}
\centering
\begin{tabular}{@{}lllllllll@{}}
\hline \hline
    \multicolumn{9}{c}{\bf LA-SiLU} \\
     & ReLU & LReLU & PReLU & SiLU & HardSiLU & Mish & GELU & ELU \\
    \hline  
    RN20 & $<1e^{-3}$ & $<1e^{-3}$ & $<1e^{-3}$ & $<1e^{-3}$ & $<1e^{-3}$ & $<1e^{-3}$ & $<1e^{-3}$ & $<1e^{-3}$ \\
    RN32 & $<1e^{-3}$ & $<1e^{-3}$ & $<1e^{-3}$ & $<1e^{-3}$ & $<1e^{-3}$ & $<1e^{-3}$ & $<1e^{-3}$ & $<1e^{-3}$ \\
    RN44 & $<1e^{-3}$ & $<1e^{-3}$ & $<1e^{-3}$ & $<1e^{-3}$ & $<1e^{-3}$ & $<1e^{-3}$ & $<1e^{-3}$ & $<1e^{-3}$ \\
    \hline  
    \\
    \hline  
    \multicolumn{9}{c}{\bf LA-HardSiLU} \\ 
     & ReLU & LReLU & PReLU & SiLU & HardSiLU & Mish & GELU & ELU \\
    \hline  
    RN20 & $<1e^{-3}$ & $<1e^{-3}$ & $<1e^{-3}$ & $<1e^{-3}$ & $<1e^{-3}$ & $<1e^{-3}$ & $<1e^{-3}$ & $<1e^{-3}$ \\
    RN32 & $<1e^{-3}$ & $<1e^{-3}$ & $<1e^{-3}$ & $<1e^{-3}$ & $<1e^{-3}$ & $<1e^{-3}$ & $<1e^{-3}$ & $<1e^{-3}$ \\
    RN44 & $<1e^{-3}$ & $<1e^{-3}$ & $<1e^{-3}$ & $<1e^{-3}$ & $<1e^{-3}$ & $<1e^{-3}$ & $<1e^{-3}$ & $<1e^{-3}$ \\
\hline \hline
\end{tabular}
\end{table}

Tables \ref{table_appendix_stat_cifar10} and \ref{table_appendix_stat_cifar100} present the results of a statistical significance test between the accuracy of networks with element-level activation functions and those with LA-SiLU functions on clean CIFAR10 and CIFAR100. Tables \ref{table_appendix_stat_cifar10c} and \ref{table_appendix_stat_cifar100c} present the corresponding results of networks with LA-SiLU and LA-HardSiLU on CIFAR10-C and CIFAR100-C. RN20, RN32, and RN44 denotes ResNet20, ResNet32, and ResNet44. We do not report the experiments of ResNet44 with PReLU on CIFAR10 and CIFAR10-C as a network exploded during training. When the accuracies of both functions were normally distributed, we performed a T-test. In cases where at least one of them are not, we performed a Wilconxon signed-rank test otherwise. The notation `$>0.05$' indicates that the $p$-value from either a T-test or a Wilcoxon signed-rank test is larger than the standard significance level of $0.05$ (i.e. $p\mbox{-value}>0.05$). 

\clearpage

%%%%%%%%%%%%%%%%%%%%%%%%%%%%%%%%%%%%%%%%%%%%%%%%%%%%%%%%%%%%%%%%%%%%%%

\section{K. Additional Figures}

\begin{figure}[ht!]
\centering
\includegraphics[width=0.48\columnwidth]{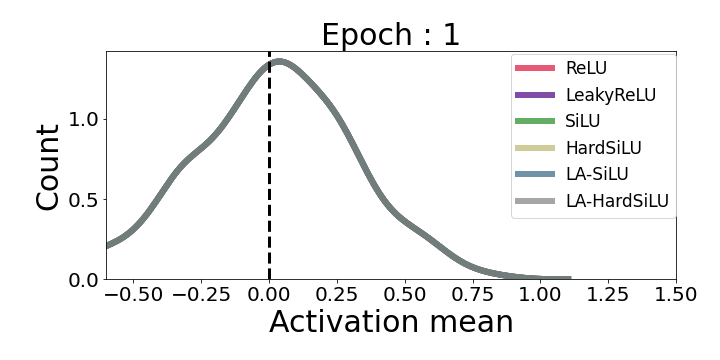} \;
\includegraphics[width=0.48\columnwidth]{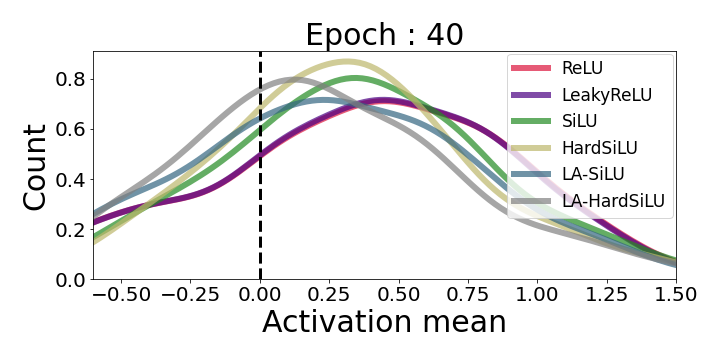} \;
\caption{Distribution of the activation \textbf{input} means of the elements in a trained network on MNIST at $1^{st}$ and $40^{th}$ epochs.}
\label{fig_MNIST_activation_mean_input}
\end{figure}

\begin{figure}[ht!]
\centering
\includegraphics[width=0.48\columnwidth]{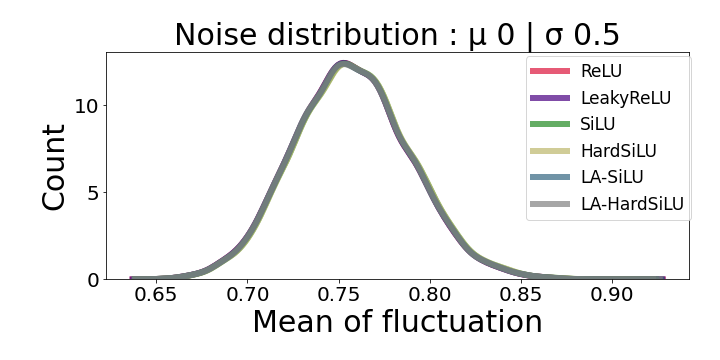}  
\includegraphics[width=0.48\columnwidth]{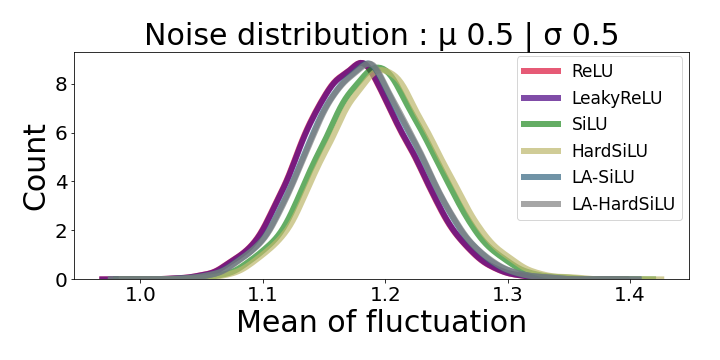} 
\caption{Distribution of activation \textbf{input} fluctuation due to noise with different noise distribution.}
\label{fig_MNIST_fluctuation_input}
\end{figure}

In this section, we present additional tables and figures extracted from the experiments. Figure \ref{fig_MNIST_activation_mean_input} presents the distribution of the mean activation input. As observed in the mean of activation input at epoch 40 (right), LayerAct functions promote the training of parameter $W$ such that the output of the linear projection $y=W^{T}x$, which is also activation input, gets closer to zero compared to other functions. This helps the activation output to exhibit a `zero-like' behaviour.

LayerAct functions exhibit a significantly lower mean and variance of activation fluctuation among the samples compared to any other element-level activation function (see Figure 3 in the main manuscript). Figure \ref{fig_MNIST_fluctuation_input} demonstrates that the distribution of mean fluctuation in activation input appears similar across all functions. This observation confirms that the lower mean and variance of activation output fluctuation of LayerAct functions is not due to a smaller fluctuation in activation input, but is a result of the inherent mechanism of LayerAct.

%%%%%%%%%%%%%%%%%%%%%%%%%%%%%%%%%%%%%%%%%%%%%%%%%%%%%%%%%%%%%%%%%%%%%%

\section{L. Medical Image}

In this section, we present the setting of experimental result from U-net \cite{ronneberger2015u} and Unet++ \cite{zhou2018unet++} for segmentation task on a nuclei image dataset from Data Science Bowl 2018 \cite{data-science-bowl-2018}. Detailed experimental setting is as follows: (1) Adam optimizer with $3e^{-4}$ learning rate, and $1e^{-4}$ weight decay, (2) training 100 epoches with cosine annealing scheduler, and (3) BCE-Dice Loss as the loss function. We report the average IoU (Intersection over Union; \%) over 10 trials with different weight initialization in Table 4 of the main manuscript.

%%%%%%%%%%%%%%%%%%%%%%%%%%%%%%%%%%%%%%%%%%%%%%%%%%%%%%%%%%%%%%%%%%%%%%

\section{M. Adversarial Robustness}

We conducted experiments to investigate the adversarial robustness of LayerAct. We utilized ReLU, SiLU, and LA-SiLU as the activation of ResNet18. The experimental setting was based on \cite{zhou2023enhancing}. The experimental results in Table 4 in the main manuscript show that the ResNet20 with LA-SiLU is more robust against adversarial attacks compared to those with other activations. These results indicate that LayerAct can enhance adversarial robustness as well. 

%% file: aaai25.bbl
\begin{thebibliography}{31}
\providecommand{\natexlab}[1]{#1}

\bibitem[{Ba, Kiros, and Hinton(2016)}]{ba2016layer}
Ba, J.~L.; Kiros, J.~R.; and Hinton, G.~E. 2016.
\newblock Layer Normalization.
\newblock arXiv:1607.06450.

\bibitem[{Clevert, Unterthiner, and Hochreiter(2016)}]{clevert2015fast}
Clevert, D.-A.; Unterthiner, T.; and Hochreiter, S. 2016.
\newblock Fast and accurate deep network learning by exponential linear units (elus).
\newblock In \emph{International Conference on Learning Representations (ICLR)}.

\bibitem[{Elfwing, Uchibe, and Doya(2018)}]{elfwing2018sigmoid}
Elfwing, S.; Uchibe, E.; and Doya, K. 2018.
\newblock Sigmoid-weighted linear units for neural network function approximation in reinforcement learning.
\newblock \emph{Neural Networks}, 107: 3--11.

\bibitem[{Glorot, Bordes, and Bengio(2011)}]{glorot2011deep}
Glorot, X.; Bordes, A.; and Bengio, Y. 2011.
\newblock Deep Sparse Rectifier Neural Networks.
\newblock In \emph{International Conference on Artificial Intelligence and Statistics (AISTATS)}, volume~15, 315--323.

\bibitem[{Goodman et~al.(2018)Goodman, Carpenter, Park, jlefman nvidia, Josette\_BoozAllen, Kyle, Maggie, Nilofer, Sedivec, and Cukierski}]{data-science-bowl-2018}
Goodman, A.; Carpenter, A.; Park, E.; jlefman nvidia; Josette\_BoozAllen; Kyle; Maggie; Nilofer; Sedivec, P.; and Cukierski, W. 2018.
\newblock 2018 Data Science Bowl.
\newblock https://kaggle.com/competitions/data-science-bowl-2018.
\newblock Accessed: 2024-12-17.

\bibitem[{He et~al.(2015)He, Zhang, Ren, and Sun}]{he2015delving}
He, K.; Zhang, X.; Ren, S.; and Sun, J. 2015.
\newblock Delving Deep into Rectifiers: Surpassing Human-Level Performance on ImageNet Classification.
\newblock In \emph{Proceedings of the IEEE International Conference on Computer Vision (ICCV)}.

\bibitem[{He et~al.(2016)He, Zhang, Ren, and Sun}]{he2016deep}
He, K.; Zhang, X.; Ren, S.; and Sun, J. 2016.
\newblock Deep Residual Learning for Image Recognition.
\newblock In \emph{Proceedings of the IEEE Conference on Computer Vision and Pattern Recognition (CVPR)}.

\bibitem[{Hendrycks and Dietterich(2019)}]{hendrycks2019robustness}
Hendrycks, D.; and Dietterich, T. 2019.
\newblock Benchmarking Neural Network Robustness to Common Corruptions and Perturbations.
\newblock In \emph{International Conference on Learning Representations (ICLR)}.

\bibitem[{Hendrycks and Gimpel(2023)}]{hendrycks2016gaussian}
Hendrycks, D.; and Gimpel, K. 2023.
\newblock Gaussian Error Linear Units (GELUs).
\newblock arXiv:1606.08415.

\bibitem[{Howard et~al.(2019)Howard, Sandler, Chu, Chen, Chen, Tan, Wang, Zhu, Pang, Vasudevan, Le, and Adam}]{Howard2019Searching}
Howard, A.; Sandler, M.; Chu, G.; Chen, L.-C.; Chen, B.; Tan, M.; Wang, W.; Zhu, Y.; Pang, R.; Vasudevan, V.; Le, Q.~V.; and Adam, H. 2019.
\newblock Searching for MobileNetV3.
\newblock In \emph{Proceedings of the IEEE/CVF International Conference on Computer Vision (ICCV)}.

\bibitem[{Huang et~al.(2018)Huang, Yang, Lang, and Deng}]{Huang2018decorr}
Huang, L.; Yang, D.; Lang, B.; and Deng, J. 2018.
\newblock Decorrelated Batch Normalization.
\newblock In \emph{Proceedings of the IEEE Conference on Computer Vision and Pattern Recognition (CVPR)}.

\bibitem[{Ioffe and Szegedy(2015)}]{ioffe2015Batch}
Ioffe, S.; and Szegedy, C. 2015.
\newblock Batch Normalization: Accelerating Deep Network Training by Reducing Internal Covariate Shift.
\newblock In \emph{International Conference on Machine Learning (ICML)}, volume~37, 448--456. PMLR.

\bibitem[{Krizhevsky(2009)}]{krizhevsky2009learning}
Krizhevsky, A. 2009.
\newblock \emph{Learning Multiple Layers of Features from Tiny Images}.
\newblock Ph.D. thesis, University of Toronto.

\bibitem[{Krizhevsky, Sutskever, and Hinton(2012)}]{krizhevsky2017ImageNet}
Krizhevsky, A.; Sutskever, I.; and Hinton, G.~E. 2012.
\newblock ImageNet Classification with Deep Convolutional Neural Networks.
\newblock In \emph{Advances in Neural Information Processing Systems (NeurIPS)}, volume~25.

\bibitem[{Labatie et~al.(2021)Labatie, Masters, Eaton-Rosen, and Luschi}]{labatie2021proxy}
Labatie, A.; Masters, D.; Eaton-Rosen, Z.; and Luschi, C. 2021.
\newblock Proxy-Normalizing Activations to Match Batch Normalization while Removing Batch Dependence.
\newblock In \emph{Advances in Neural Information Processing Systems (NeurIPS)}, volume~34, 16990--17006. Curran Associates, Inc.

\bibitem[{Lee et~al.(2015)Lee, Xie, Gallagher, Zhang, and Tu}]{lee2015deeply}
Lee, C.-Y.; Xie, S.; Gallagher, P.; Zhang, Z.; and Tu, Z. 2015.
\newblock Deeply-Supervised Nets.
\newblock In \emph{AISTATS}, volume~38, 562--570.

\bibitem[{Liang et~al.(2020)Liang, Huang, Liang, and Yang}]{liang2020instance}
Liang, S.; Huang, Z.; Liang, M.; and Yang, H. 2020.
\newblock Instance enhancement batch normalization: An adaptive regulator of batch noise.
\newblock In \emph{AAAI}, volume~34, 4819--4827.

\bibitem[{Lubana, Dick, and Tanaka(2021)}]{Lubana2021Beyond}
Lubana, E.~S.; Dick, R.; and Tanaka, H. 2021.
\newblock Beyond BatchNorm: Towards a Unified Understanding of Normalization in Deep Learning.
\newblock In \emph{Advances in Neural Information Processing Systems (NeurIPS)}, volume~34, 4778--4791. Curran Associates, Inc.

\bibitem[{Luo et~al.(2019)Luo, Zhang, Ren, Peng, and Li}]{luo2019switchable}
Luo, P.; Zhang, R.; Ren, J.; Peng, Z.; and Li, J. 2019.
\newblock Switchable normalization for learning-to-normalize deep representation.
\newblock \emph{IEEE transactions on pattern analysis and machine intelligence}, 43(2): 712--728.

\bibitem[{Maas et~al.(2013)Maas, Hannun, Ng et~al.}]{maas2013rectifier}
Maas, A.~L.; Hannun, A.~Y.; Ng, A.~Y.; et~al. 2013.
\newblock Rectifier nonlinearities improve neural network acoustic models.
\newblock In \emph{International Conference on Machine Learning (ICML)}, volume~30, 3.

\bibitem[{Misra(2020)}]{misra2020mish}
Misra, D. 2020.
\newblock Mish: A Self Regularized Non-Monotonic Activation Function.
\newblock arXiv:1908.08681.

\bibitem[{Nair and Hinton(2010)}]{nair2010rectified}
Nair, V.; and Hinton, G.~E. 2010.
\newblock Rectified linear units improve restricted boltzmann machines.
\newblock In \emph{International Conference on International Conference on Machine Learning (ICML)}, 807–814.

\bibitem[{Olaf~Ronneberger(2015)}]{ronneberger2015u}
Olaf~Ronneberger, T.~B., Philipp~Fischer. 2015.
\newblock U-net: Convolutional networks for biomedical image segmentation.
\newblock In \emph{Medical Image Computing and Computer-Assisted Intervention (MICCAI)}, 234--241.

\bibitem[{Paszke et~al.(2019)Paszke, Gross, Massa, Lerer, Bradbury, Chanan, Killeen, Lin, Gimelshein, Antiga, Desmaison, Kopf, Yang, DeVito, Raison, Tejani, Chilamkurthy, Steiner, Fang, Bai, and Chintala}]{paszke2019pytorch}
Paszke, A.; Gross, S.; Massa, F.; Lerer, A.; Bradbury, J.; Chanan, G.; Killeen, T.; Lin, Z.; Gimelshein, N.; Antiga, L.; Desmaison, A.; Kopf, A.; Yang, E.; DeVito, Z.; Raison, M.; Tejani, A.; Chilamkurthy, S.; Steiner, B.; Fang, L.; Bai, J.; and Chintala, S. 2019.
\newblock PyTorch: An Imperative Style, High-Performance Deep Learning Library.
\newblock In \emph{Advances in Neural Information Processing Systems (NeruIPS)}, volume~32.

\bibitem[{Qiu, Xu, and Cai(2018)}]{qiu2018frelu}
Qiu, S.; Xu, X.; and Cai, B. 2018.
\newblock FReLU: Flexible Rectified Linear Units for Improving Convolutional Neural Networks.
\newblock In \emph{International Conference on Pattern Recognition (ICPR)}, 1223--1228.

\bibitem[{Ramachandran, Zoph, and Le(2018)}]{ramachandran2017searching}
Ramachandran, P.; Zoph, B.; and Le, Q.~V. 2018.
\newblock Searching for activation functions.
\newblock In \emph{International Conference on Learning Representations (ICLR) Workshop}.

\bibitem[{Russakovsky et~al.(2015)Russakovsky, Deng, Su, Krause, Satheesh, Ma, Huang, Karpathy, Khosla, Bernstein et~al.}]{russakovsky2015imagenet}
Russakovsky, O.; Deng, J.; Su, H.; Krause, J.; Satheesh, S.; Ma, S.; Huang, Z.; Karpathy, A.; Khosla, A.; Bernstein, M.; et~al. 2015.
\newblock Imagenet large scale visual recognition challenge.
\newblock \emph{International Journal of Computer Vision (IJCV)}, 115(3): 211--252.

\bibitem[{Tenenbaum, de~Silva, and Langford(2000)}]{Tenenbaum2000AGlobal}
Tenenbaum, J.~B.; de~Silva, V.; and Langford, J.~C. 2000.
\newblock A Global Geometric Framework for Nonlinear Dimensionality Reduction.
\newblock \emph{Science}, 290(5500): 2319--2323.

\bibitem[{Ulyanov, Vedaldi, and Lempitsky(2017)}]{ulyanov2016instance}
Ulyanov, D.; Vedaldi, A.; and Lempitsky, V. 2017.
\newblock Instance Normalization: The Missing Ingredient for Fast Stylization.
\newblock arXiv:1607.08022.

\bibitem[{Zhou et~al.(2023)Zhou, Wang, Liu, Zhou, and Gao}]{zhou2023enhancing}
Zhou, N.; Wang, N.; Liu, D.; Zhou, D.; and Gao, X. 2023.
\newblock Enhancing Robust Representation in Adversarial Training: Alignment and Exclusion Criteria.
\newblock arXiv:2310.03358.

\bibitem[{Zhou et~al.(2018)Zhou, Rahman~Siddiquee, Tajbakhsh, and Liang}]{zhou2018unet++}
Zhou, Z.; Rahman~Siddiquee, M.~M.; Tajbakhsh, N.; and Liang, J. 2018.
\newblock Unet++: A nested u-net architecture for medical image segmentation.
\newblock In \emph{Deep Learning in Medical Image Analysis and Multimodal Learning for Clinical Decision Support (DLMIA)}, 3--11.

\end{thebibliography}


\begin{thebibliography}{10}

\bibitem{ba2016layer}
{\sc Ba, J.~L., Kiros, J.~R., and Hinton, G.~E.}
\newblock Layer normalization.
\newblock {\em arXiv:1607.06450\/} (2016).

\bibitem{Bjorck2018understanding}
{\sc Bjorck, N., Gomes, C.~P., Selman, B., and Weinberger, K.~Q.}
\newblock Understanding batch normalization.
\newblock In {\em Preceeding of NeurIPS\/} (2018), vol.~31.

\bibitem{cho2014learning}
{\sc Cho, K., Merrienboer, B., Gulcehre, C., Bougares, F., Schwenk, H., and
  Bengio, Y.}
\newblock Learning phrase representations using rnn encoder-decoder for
  statistical machine translation.
\newblock In {\em EMNLP\/} (2014), p.~1724–1734.

\bibitem{clevert2015fast}
{\sc Clevert, D.-A., Unterthiner, T., and Hochreiter, S.}
\newblock Fast and accurate deep network learning by exponential linear units
  (elus).
\newblock In {\em Preceedings of ICLR\/} (2016).

\bibitem{elfwing2018sigmoid}
{\sc Elfwing, S., Uchibe, E., and Doya, K.}
\newblock Sigmoid-weighted linear units for neural network function
  approximation in reinforcement learning.
\newblock {\em Neural Networks 107\/} (2018), 3--11.
\newblock Special issue on deep reinforcement learning.

\bibitem{Flusser2016recognition}
{\sc Flusser, J., Farokhi, S., Höschl, C., Suk, T., Zitová, B., and Pedone,
  M.}
\newblock Recognition of images degraded by gaussian blur.
\newblock {\em IEEE Transactions on Image Processing 25}, 2 (2016), 790--806.

\bibitem{hahnloser2000digital}
{\sc Hahnloser, R.~H., Sarpeshkar, R., Mahowald, M.~A., Douglas, R.~J., and
  Seung, H.~S.}
\newblock Digital selection and analogue amplification coexist in a
  cortex-inspired silicon circuit.
\newblock {\em Nature 405}, 6789 (2000), 947--951.

\bibitem{he2015delving}
{\sc He, K., Zhang, X., Ren, S., and Sun, J.}
\newblock Delving deep into rectifiers: Surpassing human-level performance on
  imagenet classification.
\newblock In {\em Proceedings of ICCV\/} (2015).

\bibitem{he2016deep}
{\sc He, K., Zhang, X., Ren, S., and Sun, J.}
\newblock Deep residual learning for image recognition.
\newblock In {\em Proceedings of CVPR\/} (2016).

\bibitem{Hochreiter1997long}
{\sc Hochreiter, S., and Schmidhuber, J.}
\newblock Long short-term memory.
\newblock {\em Neural Computation 9}, 8 (1997), 1735--1780.

\bibitem{ioffe2015Batch}
{\sc Ioffe, S., and Szegedy, C.}
\newblock Batch normalization: Accelerating deep network training by reducing
  internal covariate shift.
\newblock In {\em Proceedings of ICML\/} (2015), vol.~37, pp.~448--456.

\bibitem{krizhevsky2009learning}
{\sc Krizhevsky, A.}
\newblock Learning multiple layers of features from tiny images.
\newblock {\em Master's thesis, University of Toronto\/} (2009).

\bibitem{krizhevsky2017ImageNet}
{\sc Krizhevsky, A., Sutskever, I., and Hinton, G.~E.}
\newblock Imagenet classification with deep convolutional neural networks.
\newblock {\em Commun. ACM 60}, 6 (may 2017), 84–90.

\bibitem{labatie2021proxy}
{\sc Labatie, A., Masters, D., Eaton-Rosen, Z., and Luschi, C.}
\newblock Proxy-normalizing activations to match batch normalization while
  removing batch dependence.
\newblock In {\em Proceedings of NeurIPS\/} (2021), vol.~34, pp.~16990--17006.

\bibitem{le2007variational}
{\sc Le, T., Chartrand, R., and Asaki, T.~J.}
\newblock A variational approach to reconstructing images corrupted by poisson
  noise.
\newblock {\em Journal of mathematical imaging and vision 27\/} (2007),
  257--263.

\bibitem{lee2015deeply}
{\sc Lee, C.-Y., Xie, S., Gallagher, P., Zhang, Z., and Tu, Z.}
\newblock Deeply-supervised nets.
\newblock In {\em Proceedings of AISTATS\/} (2015), vol.~38, pp.~562--570.

\bibitem{Lubana2021Beyond}
{\sc Lubana, E.~S., Dick, R., and Tanaka, H.}
\newblock Beyond batchnorm: Towards a unified understanding of normalization in
  deep learning.
\newblock In {\em Proceedings of NeurIPS\/} (2021), vol.~34, Curran Associates,
  Inc., pp.~4778--4791.

\bibitem{maas2013rectifier}
{\sc Maas, A.~L., Hannun, A.~Y., Ng, A.~Y., et~al.}
\newblock Rectifier nonlinearities improve neural network acoustic models.
\newblock In {\em Proceedings of ICML\/} (2013), vol.~30, p.~3.

\bibitem{misra2020mish}
{\sc Misra, D.}
\newblock Mish: A self regularized non-monotonic activation function.
\newblock {\em arXiv:1908.08681\/} (2020).

\bibitem{nair2010rectified}
{\sc Nair, V., and Hinton, G.~E.}
\newblock Rectified linear units improve restricted boltzmann machines.
\newblock In {\em Proceedings of ICML\/} (2010).

\bibitem{paszke2019pytorch}
{\sc Paszke, A., Gross, S., Massa, F., Lerer, A., Bradbury, J., Chanan, G.,
  Killeen, T., Lin, Z., Gimelshein, N., Antiga, L., Desmaison, A., Kopf, A.,
  Yang, E., DeVito, Z., Raison, M., Tejani, A., Chilamkurthy, S., Steiner, B.,
  Fang, L., Bai, J., and Chintala, S.}
\newblock Pytorch: An imperative style, high-performance deep learning library.
\newblock In {\em Proceedings of NeurIPS\/} (2019), vol.~32.

\bibitem{qiu2018frelu}
{\sc Qiu, S., Xu, X., and Cai, B.}
\newblock Frelu: Flexible rectified linear units for improving convolutional
  neural networks.
\newblock In {\em Proceedings of ICPR\/} (2018), pp.~1223--1228.

\bibitem{ramachandran2017searching}
{\sc Ramachandran, P., Zoph, B., and Le, Q.~V.}
\newblock Searching for activation functions.
\newblock In {\em Proceedings of ICLR workshop\/} (2018).

\bibitem{russakovsky2015imagenet}
{\sc Russakovsky, O., Deng, J., Su, H., Krause, J., Satheesh, S., Ma, S.,
  Huang, Z., Karpathy, A., Khosla, A., Bernstein, M., et~al.}
\newblock Imagenet large scale visual recognition challenge.
\newblock {\em International journal of computer vision 115}, 3 (2015),
  211--252.

\bibitem{vijaykumar2010fast}
{\sc Vijaykumar, V., Vanathi, P., and Kanagasabapathy, P.}
\newblock Fast and efficient algorithm to remove gaussian noise in digital
  images.
\newblock {\em IAENG International Journal of Computer Science 37}, 1 (2010),
  300--302.

\bibitem{xu2016revise}
{\sc Xu, B., Huang, R., and Li, M.}
\newblock Revise saturated activation functions.
\newblock {\em arXiv:1602.05980\/} (2016).

\end{thebibliography}
